\documentclass[fleqn,10pt]{wlscirep}

\usepackage[utf8]{inputenc}
\usepackage[T1]{fontenc}
\usepackage{bm}
\usepackage{comment}
\usepackage{caption}
\usepackage{amsmath}
\usepackage{color}
\usepackage{lineno}

\usepackage[acronym]{glossaries}
\usepackage{geometry}
\usepackage{hyperref}
\usepackage[colorinlistoftodos]{todonotes}
\usepackage{comment}

\title{Noise-Aware Training of Neuromorphic Dynamic Device Networks}

\author[1,${*}$,$\dagger$]{Luca Manneschi}
\author[1,${*}$,$\dagger$]{Ian T. Vidamour}
\author[2]{Kilian D. Stenning}
\author[1]{Charles Swindells}
\author[1]{Guru Venkat}
\author[3]{David Griffin}
\author[2]{Lai Gui}
\author[2]{Daanish Sonawala}
\author[2]{Denis Donskikh}
\author[1]{Dana Hariga}
\author[3]{Susan Stepney}
\author[2]{Will R. Branford}
\author[2]{Jack C. Gartside}
\author[1]{Thomas Hayward}
\author[1]{Matthew O. A. Ellis}
\author[1]{Eleni Vasilaki}
\affil[1]{University of Sheffield, Sheffield S10 2TN, United Kingdom}
\affil[2]{Blackett Laboratory, Imperial College London, London SW7 2AZ, United Kingdom}
\affil[3]{University of York, Heslington, York YO10 5DD, United Kingdom}
\affil[*]{These authors contributed equally}
\affil[$\dagger$]{Corresponding author:  l.manneschi@sheffield.ac.uk, i.vidamour@sheffield.ac.uk}
\begin{abstract}
Physical computing has the potential to enable widespread embodied intelligence by leveraging the intrinsic dynamics of complex systems for efficient sensing, processing, and interaction\cite{markovic2020physics}. While individual devices provide basic data processing capabilities\cite{torrejon_neuromorphic_2017}, networks of interconnected devices can perform more complex and varied tasks\cite{stenning2024neuromorphic}. However, designing networks to perform dynamic tasks is challenging without physical models and accurate quantification of device noise. We propose a novel, noise-aware methodology for training device networks using Neural Stochastic Differential Equations \cite{chen2018neural,jia2019neural,tzen2019neural} (Neural-SDEs) as differentiable digital twins, accurately capturing the dynamics and associated stochasticity of devices with intrinsic memory. Our approach employs backpropagation through time\cite{werbos1990backpropagation} and cascade learning\cite{NIPS1989_69adc1e1,marquez2018deep}, allowing networks to effectively exploit the temporal properties of physical devices. We validate our method on diverse networks of spintronic devices across temporal classification and regression benchmarks. By decoupling the training of individual device models from network training, our method reduces the required training data and provides a robust framework for programming dynamical devices without relying on analytical descriptions of their dynamics.
\end{abstract}

\begin{document}
\flushbottom
\maketitle
\thispagestyle{empty}
\section{Introduction}

Before digital computation became widespread, analogue dynamical systems were key in early computational platforms, with applications ranging from solving differential equations\cite{bush1931differential} to controlling early anti-aircraft guns \cite{mindell1995automation}. These systems leveraged analogies between the inherent dynamic properties of analogue components and their target applications to replicate behaviours with a high degree of control. They were particularly valuable for real-time testing beyond the capabilities of early digital computers, trading precision for speed \cite{lundberg2005history}. However, as complementary metal-oxide semiconductor (CMOS) technology rapidly developed \cite{schaller1997moore}, digital platforms became increasingly faster and more powerful. The greater accuracy and programmability of digital computers ultimately led to the replacement of analogue systems by their digital counterparts. \\\\
More recently, the rapid expansion of machine learning has been propelled by the alignment between algorithms and hardware. Graphical\cite{jeon2021deep} processing units (GPUs) and tensor\cite{hsu2021accelerating} processing units (TPUs) have enabled the massive parallelisation of matrix operations, leading to significant performance improvements by building large models out of relatively simple computational units. However, the increased reliance on large-scale models and extensive parallelisation has also led to a worrying trend of rising energy costs \cite{strubell-etal-2019-energy}.\\\\
\emph{In-materio} computing, much like analogue computing, harnesses the natural properties of materials to perform computations, providing an efficient alternative to conventional methods for data processing \cite{markovic2020physics}. The principles of reservoir computing (RC) \cite{jaeger_echo_2001}, which originally involve fixed recurrent networks for computation, have been adapted to physical systems. In these systems, the inherent dynamics of the material serve as the computational resource \cite{appeltant_information_2011, paquot_optoelectronic_2012, torrejon_neuromorphic_2017, ababei_neuromorphic_2021, gartside2022reconfigurable}. In RC, only the output layer is trained, while the recurrent network—or, in the case of physical RC, the material—functions as a fixed temporal kernel, thus avoiding the complexities of optimising dynamic processes. However, because the internal network structure itself is not trained, achieving the desired dynamic transformations often requires a high-dimensional network, as higher dimensions increase the likelihood of finding a suitable solution. Outputs from these systems can be obtained directly from the material itself \cite{gartside_reconfigurable_2022}, through multiplexing techniques \cite{appeltant_information_2011}, or by iteratively building networks by interconnecting multiple devices based on metric evaluations \cite{stenning2024neuromorphic}. Despite these strategies, reservoir computing networks often face performance challenges when compared to networks where all parameters can be optimised using gradient-based methods. Fully optimisable networks typically perform better because they can adjust all their parameters to suit specific tasks \cite{1556090}.\\\\
To perform optimisation on \emph{in-materio} computers, general methodologies have been developed that train the interconnectivity of devices, leading to the concept of physical neural networks (PNNs) \cite{nakajima2022physical,momeni2023_PhyLL,wright2022deep}. In PNNs, each node in a neural network corresponds to a physical device. Unlike neuromorphic computing platforms designed to closely emulate biological neural architectures or systems \cite{lenk2023neuromorphic, cheng2022bioinspired, long2023neuromorphic, matrone2024modular, li2018efficient}, PNN frameworks focus on optimising the parameters that govern the interactions between devices. This approach allows for a flexible selection of material systems that offer a wide range of nonlinear responses, varying in complexity and functionality, akin to activation functions in artificial neural networks (ANNs).\\\\
Multiple approaches have emerged for optimising PNNs. The Physics Aware Training (PAT) method \cite{wright2022deep} involves measuring device responses and estimating derivatives for backpropagation using a digital twin—a faithful model of the device. More recently, methods that avoid digital twins have been developed, using direct feedback alignment \cite{nokland2016direct, nakajima2022physical} or forward-forward algorithms \cite{hinton2022forward, momeni2023_PhyLL} to optimise without gradient backpropagation. These methods approximate gradient descent with techniques directly applicable to the physical substrate, where devices provide simple transfer functions on current inputs. However, no existing approach can optimise PNNs in systems with dynamic behaviours and intrinsic memory—memory due to the inherent properties of the device materials—in general settings. Current methods assume devices are static and memoryless, and thus cannot optimise or leverage dynamic processes. As a result, they are unable to utilise functional memory sources, which are essential for temporally-driven tasks and \textit{in-memory} computation. To fully harness material computational capabilities, a device-agnostic optimisation method that accounts for dynamic processes is needed. \\\\
An important initial step in this direction was made with the proposal of using neural ordinary differential equations\cite{chen2018neural} (neural-ODEs)  to model dynamic devices, with their feasibility demonstrated through simulations \cite{chen2022forecasting}. However, these models are not capable of capturing the noise in the system, which we hypothesise is essential for the robust transferability of parameters that control the interactions among devices from simulation to physical dynamical devices.\\\\
In this paper, we present a universal framework for gradient-based optimisation in deep networks of interacting dynamical systems. Our method does not require a mathematical description of the physical system, is entirely data-driven, and can be applied to any device that can be modelled as a differential equation, as long as sufficient sampling of input-output relationships is possible. To achieve this, we develop a generalised formulation of neural stochastic differential equations (Neural-SDEs) \cite{jia2019neural, tzen2019neural, kidger2022neural} capable of capturing coloured noise, where different frequencies have varying power levels in the power spectral density, representing realistic noise characteristics observed in physical devices.\\\\
We apply our methodology to experimental spintronic devices previously used in neuromorphic computing applications \cite{dawidek2021dynamically, vidamour2022quantifying, vidamour2023reconfigurable, venkat2024exploring, gartside2022reconfigurable, stenning2024neuromorphic}. We demonstrate that noise modelling is crucial for transferring performance from simulations to networks of devices, allowing us to achieve high accuracy in classification and regression tasks, including a neuroprosthetic task, for the first time in fully optimised dynamic PNNs. Additionally, by employing cascade learning \cite{NIPS1989_69adc1e1, marquez2018deep}—building the network layer by layer—we illustrate that, in principle, this methodology could be extended to arbitrarily deep networks, requiring only limited experimental data for each layer. This work marks a significant advancement in the application of complex material systems to PNNs, enabling gradient-descent-based, noise-aware optimisation of the connectivity of arbitrary, mathematically-agnostic devices with intrinsic memory.
\section{Results}
\begin{figure*}[t!]
\includegraphics[width=0.99\columnwidth]{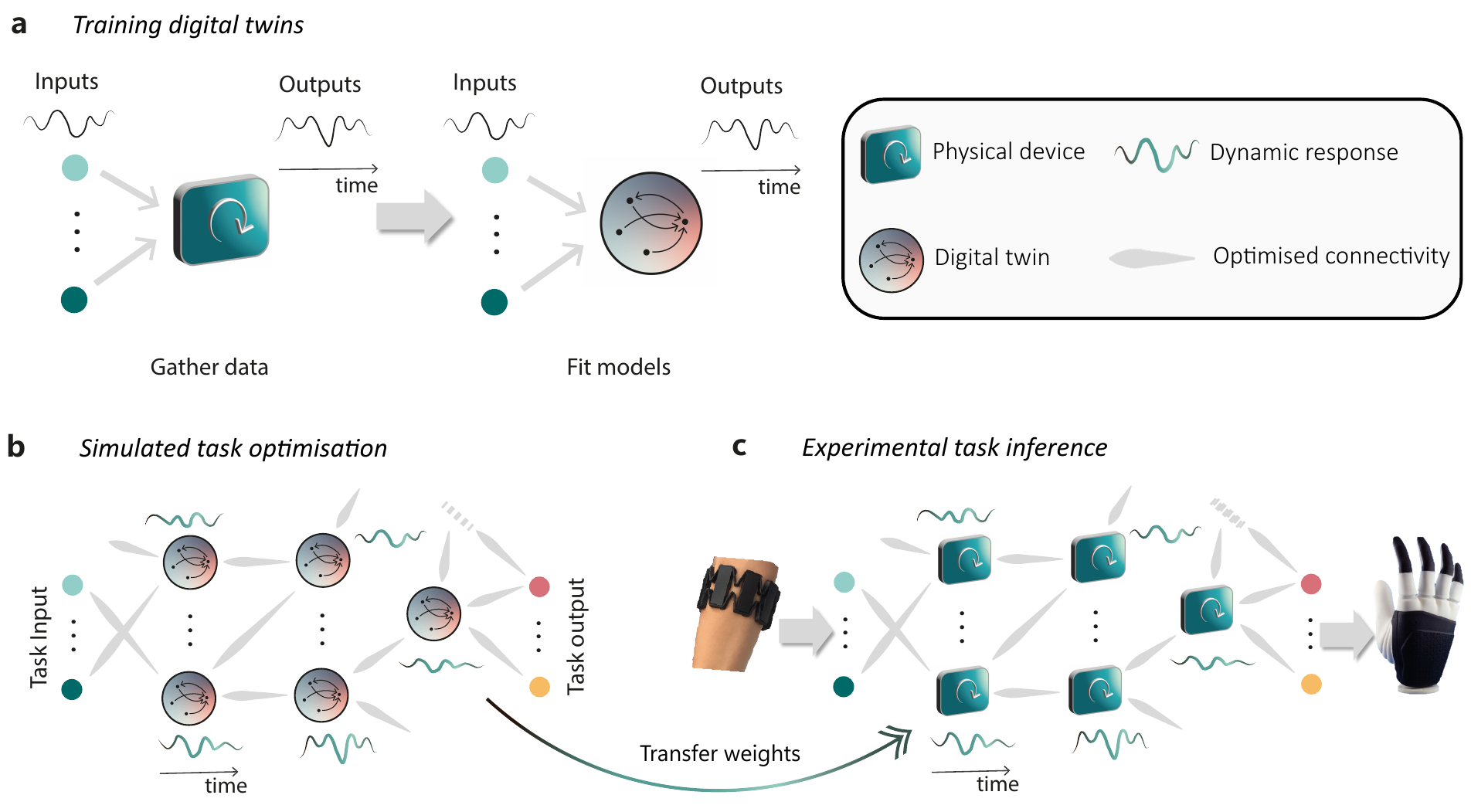}
\centering
\caption{\textbf{Overview of the dynamical network optimisation framework.}\small{ (\textbf{a}) Model Generation: Experimental devices are driven under random inputs, their observable states are recorded, and these data are used to fit models of device dynamics. (\textbf{b}) Network Simulation: A neural network is constructed where each node replicates the dynamics of the original device, using the trained model. Parameters controlling device interactions (network weights) are optimised for a task via backpropagation through time (BPTT) or truncated-BPTT on the interacting digital twins. (\textbf{c}) Experimental Transfer: The parameters optimised in simulation are transferred like-for-like to experimental networks where each node is a real device, and task performance evaluated}  }     
\label{fig:Fig1_scheme.}
\end{figure*}

The optimisation process for networks of arbitrary dynamical devices involves three distinct phases, as illustrated in Figure \ref{fig:Fig1_scheme.}. First, differentiable digital twins—models that allow for the calculation of derivatives using standard tools—are trained to replicate the input-output responses of devices based on experimentally collected data (Figure \ref{fig:Fig1_scheme.}a). Next, these digital twins are used in network simulations of devices, where the interactions between the devices are optimised (Figure \ref{fig:Fig1_scheme.}b). Finally, the optimised parameters from the simulations are transferred directly to the physical network, where performance in benchmark tasks is assessed (Figure \ref{fig:Fig1_scheme.}c). An overview of these stages is provided below, with more detailed explanations available in the supplementary material.

\subsection*{Neural-SDEs as differentiable digital twins}
\begin{figure*}[h!]
\includegraphics[width=0.99\columnwidth]{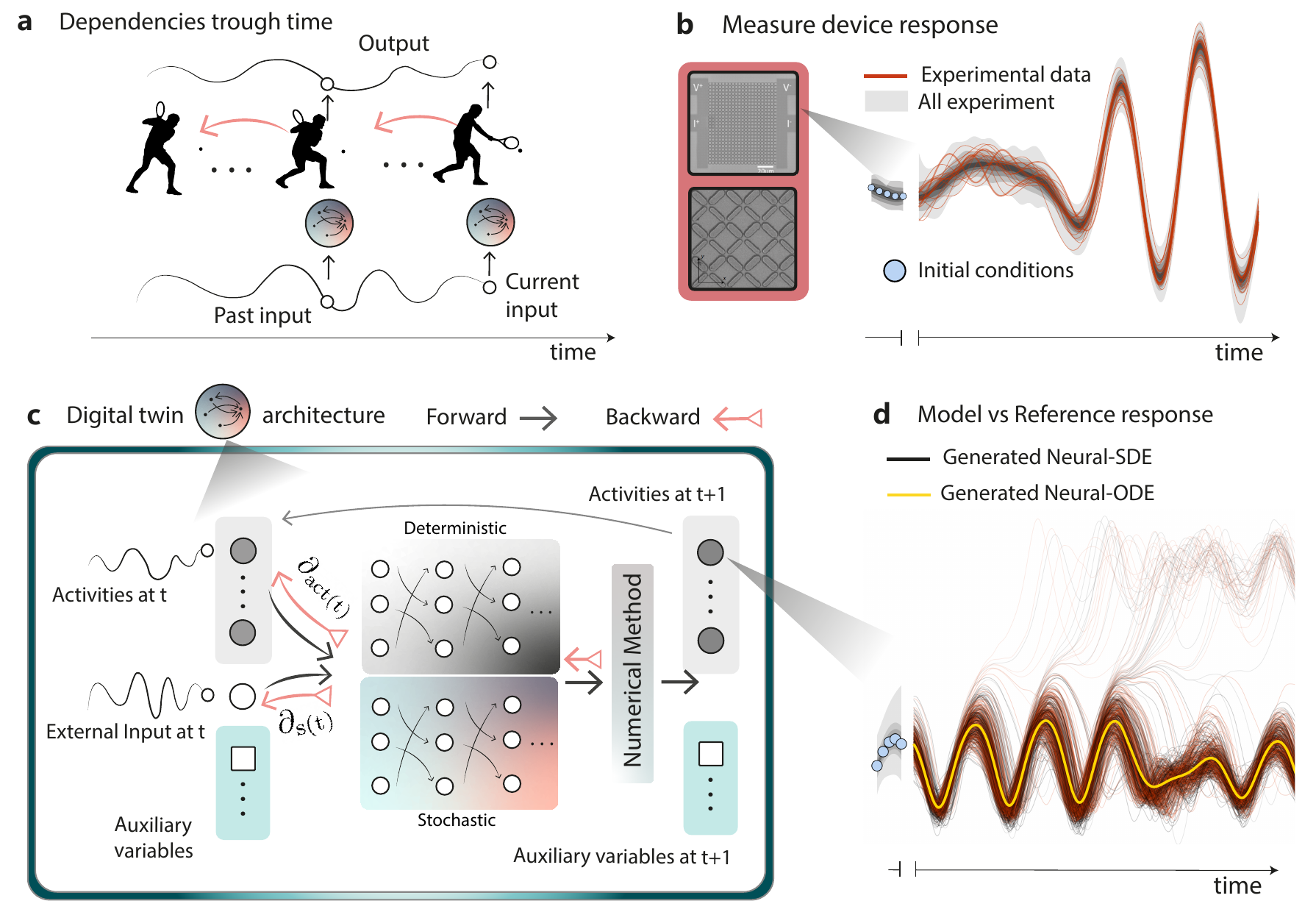}
\centering
\caption{\textbf{Modelling and optimising dynamic behaviours.} \small{} \textbf{(a)} Schematic analogy of temporal dependencies. Altering an action in the past has consequences for all future actions. Similarly, for backpropagation through time, changes to the final output caused by all past inputs and states must be taken into consideration. \textbf{(b)} Schematic showing samples from distributions of initial conditions, which subsequently affect the predicted trajectory. Grey clouds show the distribution of all gathered data for a given random input sequence, while red lines highlight specific trajectories. \textbf{(c)} Schematic diagram of the Neural-SDE architecture. Inputs of device states (activities), external driving stimuli, and auxiliary variables feed into a pair of distinct neural networks that handle the deterministic (upper network) and stochastic (lower network) behaviours. The output of these networks feeds into a numerical ODE solver, generating predictions of both activities and auxiliary variables for the next timestep. The results are recursively fed back as inputs to the next timestep prediction, generating predicted trajectories from initial conditions and external driving signals. Black arrows show forward propagation of activities; orange arrows show backward propagation of gradients. \textbf{(d)} Comparison between predictions generated via neural-ODE and neural-SDE models. The neural-ODE produces a single deterministic outcome for a given set of initial conditions and input stimuli, shown by the yellow line. The neural-SDE instead generates sampled trajectories from a distribution based on the learned noise characteristics. The black lines show 100 generations of a signal via the neural-SDE, while red lines show real experimental data from repeated identical input sequences. As in (b), blue circles represent selected initial conditions and the grey clouds represent the distributions observed across all experiments.}     
\label{fig:Fig2_scheme}
\end{figure*}

Previous work on fully-optimised PNNs has focused on devices without intrinsic memory. In contrast, networks of devices with intrinsic memory significantly increase training complexity because past inputs and states directly influence current behaviour. This is analogous to a tennis player trying to hit a moving ball, as shown schematically in Figure \ref{fig:Fig2_scheme}(a). Any change in the position of the ball or the player affects all subsequent actions and movements. Similarly, in dynamical systems, adjustments made at any point in time can propagate through the network, affecting future states and complicating optimisation. Algorithms that account for dynamic behaviours, such as backpropagation through time (BPTT) \cite{werbos1990backpropagation} and more recent methods \cite{bellec2019biologically, kag2021training}, are crucial. These algorithms capture the influence of past changes on future states, enabling optimisation over temporal sequences. However, when using BPTT with digital twins, mismatches between the model and experimental data can accumulate over time, especially in deeper networks, where errors compound with each additional operation.
In addition, physical systems rarely exhibit perfectly deterministic responses. Experimental noise cannot be encapsulated by deterministic models, and the use of noise-free models in the optimisation process leads to sub-optimal solutions when transferred to noisy environments. By characterising noise in the simulation process, we further close the simulation-reality gap and enable noise-tolerant network structures to be discovered, improving task-based performance of transferred networks.\\\\
In this work we consider two spintronic systems, nano-magnetic ring arrays (NRA)\cite{dawidek_dynamically-driven_2021,vidamour2022quantifying,vidamour2023reconfigurable,venkat2024exploring} and artificial spin vortex ice (ASVI)\cite{gartside2022reconfigurable,stenning2024neuromorphic}. As shown in Figure~\ref{fig:Fig2_scheme}(b) for the NRA, the measured device responses vary over different repetitions of the same input sequence, with different distributions of response depending upon current and past inputs. The stochasticity arises from intrinsic physical dynamics and from noise in experimental measurements. To incorporate the dynamic and stochastic nature of these devices into learning networks we have employed neural-SDEs as digital twins, enabling realistic output and gradient predictions within the dynamic learning framework. More detailed analysis of the models and their resulting fits can be found in the supplementary material.
Figure \ref{fig:Fig2_scheme}(c)  illustrates the network architecture of the proposed Neural-SDE model, which consists of two distinct neural networks designed to capture deterministic and stochastic dynamics, respectively. The deterministic network (upper) and the stochastic network (lower) both receive external input signals and past device states. Additionally, the stochastic network is provided with auxiliary variables to assist in modelling noise. The outputs from these networks are processed through a stochastic numerical integration scheme to compute the device’s activity (i.e., device readout) for the next timestep. This activity is recursively fed back into the model as the device’s most recent state. The orange arrows indicate the error gradients with respect to device activity and external input, calculated during BPTT. 

\noindent Figure \ref{fig:Fig2_scheme}(d) compares the outputs of the two models. The neural-ODE model provides deterministic predictions that capture the main trend for given inputs. In contrast, the neural-SDE model generates a range of trajectories that approximate the observed distributions for the same inputs. The reference dynamics (red lines) show noise-induced bifurcations, which are not captured by the neural-ODE model but are effectively modelled by the neural-SDE model.

\begin{figure*}[h!]
\includegraphics[width=1\columnwidth]{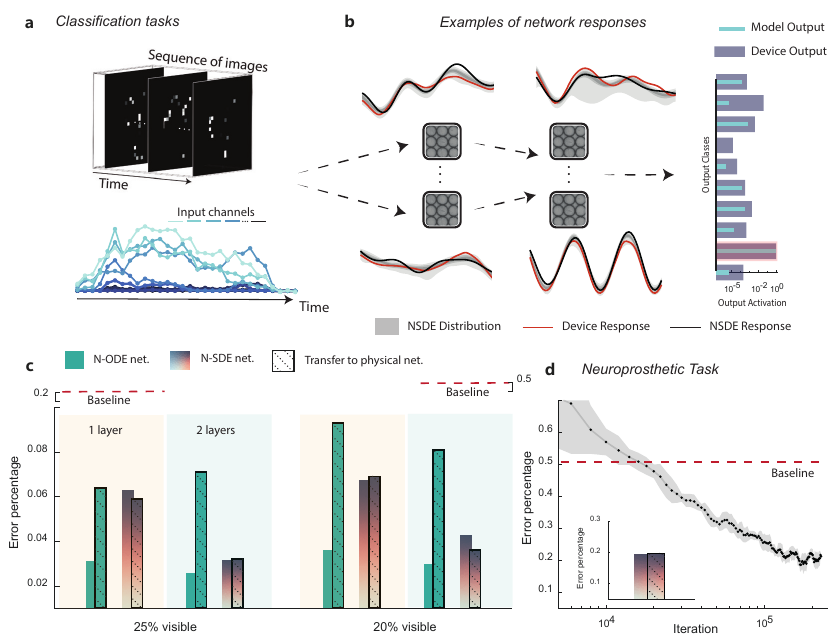}
\caption{\textbf{Partially-observable MNIST and Neuroprosthetic movement classification tasks.} \small{ (\textbf{a}) The MNIST data, presented as sequences of images, have been adapted into a temporal problem by partially obscuring the images at each time step, requiring the system to integrate information over time for accurate classification. The neuroprosthetic gesture recognition task is characterized by input channels that vary over time. (\textbf{b})  Example responses from the network’s physical nodes, showing experimentally measured responses (red) and digital twins’ responses (black) for different nodes across two layers. The gray areas represent the distribution of responses from the digital twins, while the dashed arrows illustrate the flow of information from the input through the layers to the output. The horizontal bars indicate the output activations of different physical devices representing classes compared to the model output, with the correct class highlighted in red.
 (\textbf{c}) Transferred performance of nanoring array networks using Neural-ODEs and Neural-SDEs as digital twins in the MNIST benchmark. The deterministic Neural-ODE models exhibit unrealistically high performance in simulation, which significantly deteriorates in experiments. In contrast, the noise-aware training provided by Neural-SDEs maintains high performance on physical devices, demonstrating effective exploitation of node dynamics and robustness during device transfer. (\textbf{d}) Performance of the Neural-SDE models on neuroprosthetic gesture recognition, demonstrating the framework’s potential in addressing real-world tasks. The black line represents the error as a percentage across iterations. The inset shows the final performance, comparing simulation results with those after transfer to the physical device.} }
\label{fig:Fig3}
\end{figure*}
\subsection*{Temporal classification benchmarks}
Networks of interacting devices were trained using digital twin models to perform a classification task on a modified version of the MNIST dataset. To introduce a memory component, each MNIST digit image was split into n separate images containing a random subset of the original pixels. When combined, these images reconstruct the full digit. The partial images were presented to the physical system sequentially, requiring it to use memory to classify the original digit from the sequence. Final predictions were based on the network’s response at the end of the sequence. As an additional, more challenging task, the NRA networks were trained to recognise gestures for a neuroprosthetic device \cite{atzori2014electromyography}. This task used real-world electromyography data collected from the forearms of patients performing seventeen different hand and wrist movements. Predictions were made based on the class with the highest output within a window representing data acquired between 120–180 ms after the gesture began (see \ref{fig:Supp_pNeuro}).
\\\\ 
Figure \ref{fig:Fig3}(a) provides an illustration of the tasks. Figure \ref{fig:Fig3}(b) compares the responses of the physical network of magnetic nanorings to their simulated counterparts. The dynamics of four nodes from two different layers are shown in red. Simulated activities are in black, with the digital twins’ response distribution in gray. The horizontal bars represent output activations of different physical devices for a given input frame. Despite some mismatches, the simulated network’s activities correlate well with the experiment, leading to the same classification outcome. Figure \ref{fig:Fig3}(c) displays the predicted and experimentally achieved accuracies for different task difficulties, measured as the percentage of visible pixels, in networks with single and two hidden layers. The task proved challenging for networks with a single hidden layer due to the well-established non-linearity/memory trade-off \cite{inubushi2017reservoir}, where a single hidden layer is tasked with both remembering past inputs and non-linearly combining the information simultaneously. This challenge is also observed in analytical systems, where performance significantly improves when the network architecture includes more than one layer (see Supplementary Information). \\\\
To demonstrate the importance of noise awareness in network optimisation, we compare Neural-ODEs and Neural-SDEs as digital twins (Figure \ref{fig:Fig3}(c)). Neural-ODEs, which do not incorporate noise, fail to provide information about noisy response regions that should be avoided during optimisation. This limitation results in inaccurate predictions of the physical neural network’s performance, particularly when additional hidden layers are added. In contrast, Neural-SDEs account for noise, enabling the optimisation process to identify parameter values that remain robust under stochastic variations and experimental conditions. These findings underscore the necessity of incorporating noise into digital twins to achieve reliable network optimisation. The baseline performance, indicated by red dashed lines, represents the accuracies of physical neural networks with identical architectures but randomised connectivity, analogous to the reservoir computing paradigm, for both one-layer (left, orange) and two-layer structures (right, green). Here, weights are randomly sampled from distributions similar to those of the Neural-SDE-optimised network weights (see Supplementary Information). This randomisation results in notably low performance, particularly when it fails to produce meaningful inputs for the second layer. These observations underscore the advantages of optimised connectivity using our framework for physical systems.\\\\
The two-layer networks of nanorings optimised through interacting neural-SDEs demonstrate accurate transfer to devices and achieve performance comparable to software-based MLPs, which have complete access to the input image at any given time and a similar number of trainable parameters. This indicates that the magnetic nanoring PNN effectively leverages the intrinsic dynamics of the physical nodes as memory sources, highlighting the proposed framework’s capability to utilise physical dynamics for in-memory computing.\\\\
The optimisation procedure was also applied to the neuroprosthetics task. Figure \ref{fig:Fig3}(d) shows the experimentally measured performance of a network with connectivity values transferred from optimisations performed on a simulated network using digital twins of the devices. As a baseline, we used a randomly connected network where only the output layer was trained, similar to standard reservoir computing (RC), with the same number of network nodes as the PNN. The optimised network demonstrates a reduction in error rate by approximately 0.3 compared to the baseline, indicating the improved performance of optimised connectivity over standard \emph{in-materio} RC.

\begin{figure*}[hb!]
\includegraphics[width=1\columnwidth]{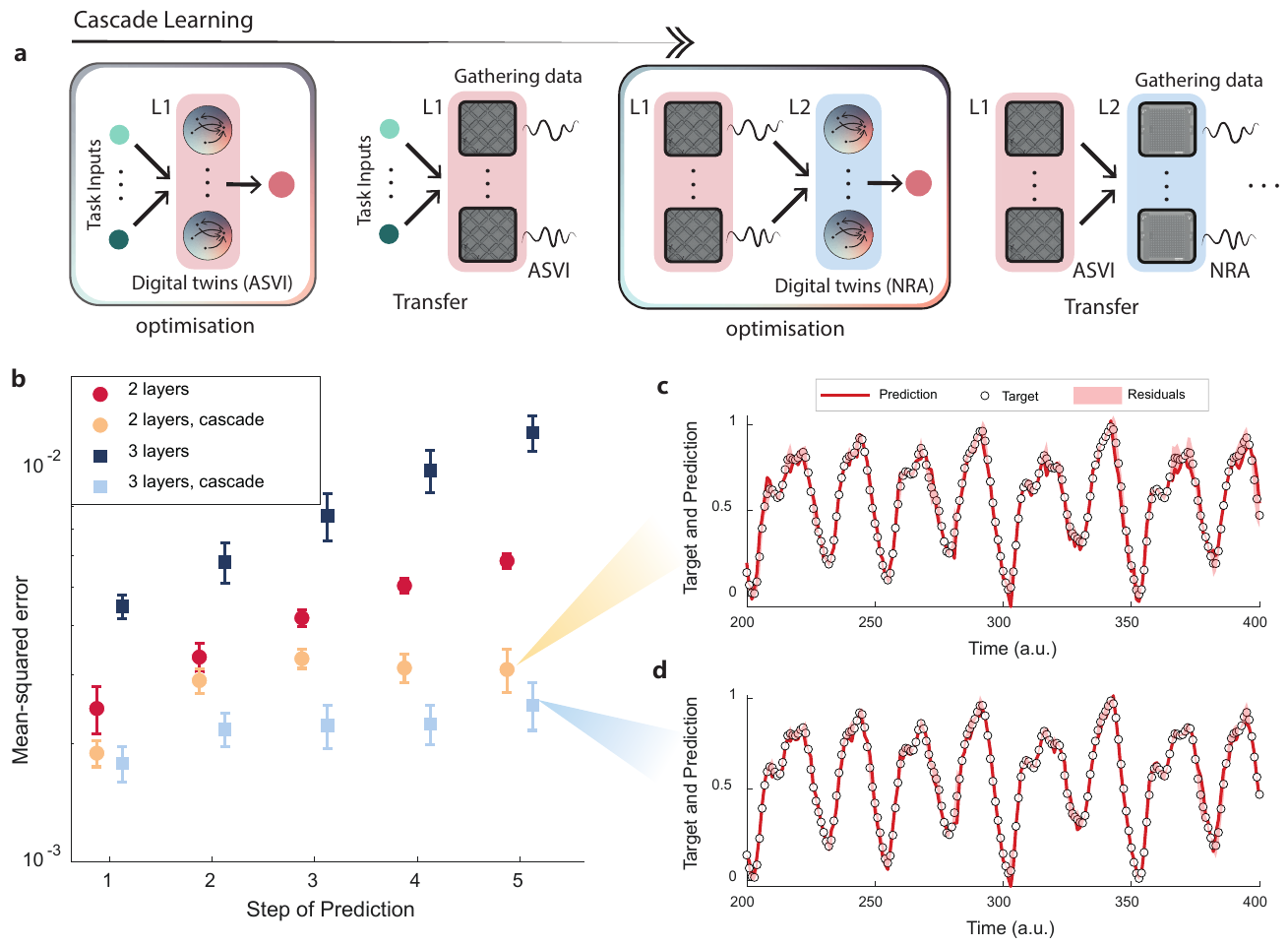}
\caption{\textbf{Cascade learning and Mackey--Glass future prediction task.} \small{ (\textbf{a}) Schematic overview of the methodology employed for sequentially training network layers with intermediate data gathering. The boxes represent steps performed in simulations, with red shading indicating ASVI twins/experiments in the first layer (L1) and blue shading representing NRAs in the second layer (L2). Initially, a single ASVI layer is connected to a simulated output neuron and trained for the regression task. Once trained, the connectivity from the input to the ASVI layer is transferred to the physical device. Experimental data is then collected to serve as input for training the connectivity to the subsequent layer, consisting of NRAs. This process can, in principle, be extended to accommodate any number of layers. Retraining the digital twin is not required; intermediate data are used solely to adjust the connectivity between the new and the previous layer. \textbf{(b)} Mean-squared error between ground truth and experimental network predictions for the Mackey–Glass future prediction task as the number of future steps increases. Circles/squares represent networks with two/three hidden layers, while dark/light colors compare direct training of the entire network to networks trained using cascade learning, as presented in panel (a). Comparison between model prediction and ground-truth data for the five-timestep future prediction of the Mackey–Glass equation in \textbf{(c)} two-layer and \textbf{(d)} three-layer networks. White circles represent the ground truth data, red lines show the transferred PNN prediction, and pink shading indicates the difference between the ground truth and the network prediction.}}
\label{fig:Fig5}
\end{figure*}

\subsection*{Regression Benchmark}
Classification problems are generally more forgiving when transferring parameters, as their winner-takes-all algorithm only requires the highest activation in the correct class for accurate prediction. In contrast, regression problems demand specific, continuous output values from the network, presenting a more challenging task. To test the limits of our methodology, we applied the network to predict the Mackey–Glass system operating in a regime characterised by quasi-periodic, chaotic behaviours. This network comprised a mixture of NRA and ASVI nodes. The network structure was designed to leverage the distinct characteristics of each physical system: the first layer featured ASVIs, which exploited their high output dimensionality to project the low-dimensional Mackey–Glass signal into a higher-dimensional space, while the subsequent layers consisted of NRAs to provide nonlinearity and memory for learning the underlying dynamics.\\\\
To mitigate error accumulation with an increasing number of layers, which is more pronounced in regression problems, we employed the cascade-correlation algorithm \cite{NIPS1989_69adc1e1}, adapted for experimental settings. A schematic of this process is shown in Figure \ref{fig:Fig5}(a).
In this approach, hidden layers were trained sequentially, with previously trained parameters remaining fixed. As a result, the learning process treats the responses of earlier layers as fixed inputs for the layer currently undergoing optimisation. Once a layer was optimised, experimental node activities were gathered using the learned parameters, generating ‘ground-truth’ data with zero mismatch in the forward pass for the subsequent layer to be trained. This strategy helped to correct the digital twin’s simulation-reality gap, which might otherwise be amplified throughout the network depth. Cascade-correlation, therefore, limits the propagation of errors to a single layer, facilitating better transfer. In this respect, the methodology is similar to Physics-Aware Training\cite{wright2022deep}, but requires only one epoch of data per device in the physical neural network structure, rather than continuous sampling during each iteration (see Methods).\\\\
Figure \ref{fig:Fig5}(b) shows the mean squared errors between the ground truth dynamical equations and predictions for increasing steps into the future, using transferred networks with two (red) and three (blue) hidden layers. Cascade learning achieves the best performance, demonstrating excellent alignment between the target and prediction, as shown in Figure \ref{fig:Fig5}(c). Without the corrective dataset, performance deteriorates as the network depth increases. However, retraining with the corrective dataset between layers reduces overall error, and adding more layers improves performance. This outcome highlights the scalability of the methodology, enabling the construction of deeper networks while minimising additional data collection. By confining mismatch error to a single layer, this approach can be extended to create arbitrarily deep dynamic PPNs.
\section{Discussion}\label{sec3}
Devices with complex dynamical responses are powerful substrates for the physical implementations of neural networks tasked with temporal processing. While individual devices may possess limited computational properties, the learned connections in device networks bring them closer to achieving the performance levels of deep artificial neural networks. This work has highlighted the critical role of a noise-aware training framework in optimising connectivity within dynamical physical neural networks. Central to this framework is the development of stochastic digital twins based on the neural-SDE approach. These models are differentiable and provide a surrogate gradient for optimising network connectivity through gradient-based methods tailored to specific tasks. Notably, our framework requires no prior knowledge of the systems under consideration, and minimises the use of the physical device, expediting training when data gathering is slow. We have demonstrated the effectiveness of this framework by successfully training networks of complex physical neurons to solve a range of temporal tasks: partially observable MNIST classification, forward prediction of the Mackey--Glass sequence, and gesture recognition for a neuroprosthetic device. \\\\
Previous methods for training physical networks, such as the Physics Aware Training (PAT)\cite{wright_deep_2022} and Physical Local Learning (PhyLL)\cite{momeni2023_PhyLL}, were limited to static devices. Our framework, in contrast, embraces the dynamical nature of the physical systems, leveraging this complexity as a computational resource rather than viewing it as an obstacle to optimisation. Furthermore, adopting neural-SDEs as the physical model provides a robust approach for generating noisy samples during the digital training. This is crucial for ensuring a successful transfer of the optimised connectivity from the digital twin to the physical hardware, mitigating the accumulation of errors that can occur throughout the network structure.\\\\
For the first time, we demonstrate the success of neural-SDEs,  extended to include coloured noise, on experimental data gathered from neuromorphic systems. Our framework is designed for generalisation: it is applicable to any device where its dynamical responses, both deterministic and stochastic, can be sampled. While previous works have demonstrated the potential of neural-ODEs on spintronic simulations\cite{chen2022forecasting}, here, we have applied this approach to two distinct experimental devices, nanomagnetic ring arrays and artificial spin vortex ice, with a particular focus on capturing their stochastic behaviours.
Finally, this work establishes a universal optimisation framework that not only optimises external inputs but also allows tuning of device hyper-parameters, enabling precise control of their operational regimes within the physical neural network. This approach is not limited to neuromorphic computing, but can be broadly applied across various fields, including robotics, neuroscience, and other scientific domains where the optimisation of dynamic processes for a given goal is critical.

\section{Methods}
\subsection*{Sampling Device Behaviours}
First, the range of inputs at which the devices are dynamically active was established by sweeping input stimuli and observing changes in measured output. Data used for training the models of dynamic behaviours were sampled randomly from the determined input range. Different datasets were constructed for training the deterministic model and stochastic model behaviours. In both cases, the systems are initialised by a strong pulse of magnetic field, saturating the devices. A single input corresponding to the maximum allowed input value is then applied, generating a trajectory to be used for initial conditions of the model. For the Neural-ODE, devices were then driven by many uncorrelated, randomly generated input sequences sampled from a uniform distribution spanning the range of activity, with the measured state of the devices recorded alongside the external input at each time. To gather a validation set for the optimisation of the neural-SDE, the devices were driven by 100 repetitions of each sequence from a smaller set of randomly sampled sequences, with similar recordings of input and measured state.
\subsection*{Neural-ODE Modelling}
The Neural-ODE models used here emulate the observable state of a dynamical system $\mathbf{x}(t)$, where $\mathbf{x}(t)$ is an $\rm{N}_x$-dimensional vector gathered experimentally. This is done by parameterising the instantaneous gradient of the dynamical systems with respect to its current hidden state $\mathbf{y}(t)$ and external input $\mathbf{s}(t)$ via a neural network $\mathbf{f}$, before integrating to find the next state. This process is described in further detail below, with a didactic tutorial provided in the supplementary material. \\\\
As in \cite{chen2022forecasting}, the hidden state of the system considered is assumed to be well represented by concatenating a set of delayed measurements $\mathbf{x}(t)$ to $\mathbf{x}(t-\rm{N}_{delay}\delta t)$, where $\rm{N}_{delay}$ is the number of delays adopted. We define this augmented $\rm{N}_x\big( \rm{N}_{delay}+1\big)$-dimensional state as $\mathbf{y}(t)=\Big(\mathbf{x}(t),...,\mathbf{x}(t-\rm{N}_{delay}\delta t)\big)$. Assuming this representation renders the system Markovian for a given $\mathbf{s}(t)$ and $\mathbf{y}(t)$, device dynamics can be sampled via a dataset $\mathcal{D}=\Big\{ \big( \mathbf{s}(0), \mathbf{y}(0), \mathbf{y}(1) \big), ..., \big( \mathbf{s}(t), \mathbf{y}(t), \mathbf{y}(t+\delta t) \big), ... \Big\}$ described above, with $\mathbf{f}$ acting to interpolate between points in the gathered dataset after training. \\\\
To predict trajectories, the Neural-ODE is provided with initial conditions $\mathbf{y}(t_0)$ sampled from a random starting time $t_0$. Then, driven by external signals $\Big(\mathbf{s}(t_0),, ..., \mathbf{s}(t_0+\rm{T}\delta t)\Big)$, the model is asked to predict the evolution $\Big(\mathbf{y}(t_0+\delta t), ... , \mathbf{y}(t_0+\rm{T} \delta t)\Big)$ of the system for $\rm{T}$ steps. We will denote the predicted activities generated by the model as $\tilde{\mathbf{y}}(t)$ to differentiate them from the target $\mathbf{y}(t)$. The neural-ODE activities are defined through the integration of the following 
\begin{align}
    \rm{d}\tilde{\mathbf{y}}(t)=\mathbf{f}\big(\tilde{\mathbf{y}}(t),\mathbf{s}(t),t|\phi^{f}\big)\rm{d}t \\
    \mathbf{\tilde{y}}(t_0)=\mathbf{y}(t_0) \nonumber
\end{align}
At each timestep, the neural network $\mathbf{f}(\cdot|\phi^{f}):\mathcal{R}^{\rm{N_x}(\rm{N}_{delay}+1)+\rm{N}_s}\rightarrow \mathcal{R}^{\rm{N_x}(\rm{N}_{delay}+1)}$, parameterised by weights $\phi^{f}$, estimates the instantaneous gradient of the system from $\tilde{\mathbf{y}}(t)$ and $\mathbf{s}(t)$. Integration via numerical methods leads to the prediction of $\mathbf{x}(t+\delta t)$ as the most recent state in $\tilde{\mathbf{y}}(t+\delta t)$. This iterative, recursive process continues for all steps considered $t \in \big[t_0, \ t_0+\rm{T}\delta t \big]$. The network is trained by minimising the mean squared error between model generated states $\hat{\mathbf{y}}$ and experimentally gathered states $\mathbf{y}$. We refer to the Supplementary Material for more details.
\subsection*{Neural-SDE Modelling}
The Neural-SDE model proposed captures both the stochastic \cite{kidger2021neural} and deterministic behaviours of dynamical systems, as well as noise in experimental measurements. The model features an additional network $\mathbf{g}$ to the Neural-ODE model introduced in the previous section. To account for various experimental settings, the neural-SDE has been designed to accommodate external signals, delayed observations of systems activities, and the presence of coloured noise.
The incorporation of coloured noise involves the introduction of $\rm{N}_a$ auxiliary variables $\mathbf{a}(t)$ operating over multiple timescales. The equations defining the proposed neural-SDE are: 

\begin{align} \label{Eq.Neural_SDE_Methods1}
    \begin{pmatrix} \rm{d}\tilde{\mathbf{y}}(t) \\  \rm{d}\mathbf{a}(t) \end{pmatrix} & = \begin{pmatrix} \mathbf{f}(\tilde{\mathbf{y}}(t),\mathbf{a}(t),\mathbf{s}(t),t|\phi^{f}) \\ -\mathbf{a}(t)\odot1/\boldsymbol{\tau}\end{pmatrix}\rm{d}t+\mathbf{g}(\tilde{\mathbf{y}}(t),\mathbf{a}(t),\mathbf{s}(t),t|\phi^{g}) \rm{d}\mathbf{W} \\
    \tilde{\mathbf{y}}(t_0) & =\mathbf{y}(t_0) \\
    \mathbf{a}(t_0)& =\mathbf{0}    
\end{align}
where $\rm{d}\mathbf{W}$ is a $\rm{N}_{W}$-dimensional Wiener process \cite{kloeden1992stochastic}; the functions $\mathbf{f}(\cdot|\phi^{f})$ and  $\mathbf{g}(\cdot|\phi^{g})$ are neural networks defined by the trainable weights $\phi^{f}$ and $\phi^{g}$ respectively; $\boldsymbol{\tau}$ is a vector defining the multiple timescales of accumulation of information of the auxiliary variables $\mathbf{a}(t)$, $\odot$ refers to element-wise multiplication, and $1/\boldsymbol{\tau}$ is the element-wise reciprocal of $\boldsymbol{\tau}$. 
In this case, we generally found that it is important to constrain the function $\mathbf{g}$ to generate stochasticity only on the most recent prediction of $\mathbf{\tilde{y}}$. Expanding the augmented $\mathbf{y}(t)$ into the measurements $\mathbf{x}(t)$ and introducing the constraints on the $\mathbf{g}$ function leads to rewrite Eq. \ref{Eq.Neural_SDE_Methods1} as

\begin{equation} \label{Eq.N-SDE_delayed_0aux}
    \begin{pmatrix} \rm{d}\tilde{\mathbf{x}}(t) \\ \rm{d}\tilde{\mathbf{x}}(t-\delta t) \\ \vdots \\ \rm{d}\tilde{\mathbf{x}}(t-\rm{N}_{delay}\delta t) \\ \rm{d}\mathbf{a}(t)\end{pmatrix}=
    \begin{pmatrix} \mathbf{f}_1(\tilde{\mathbf{y}}(t),\mathbf{s}(t),t|\phi^{f})+\mathbf{A}\mathbf{a}(t) \\
    \mathbf{f}_2(\tilde{\mathbf{y}}(t),\mathbf{s}(t),t|\phi^{f}) \\ \vdots \\ 
    \mathbf{f}_{\rm{N}_{delay}+1}(\tilde{\mathbf{y}}(t),\mathbf{s}(t),t|\phi^{f}) \\ - \mathbf{a}(t) \odot \dfrac{1}{\boldsymbol{\tau}} \end{pmatrix}\rm{d}t+
     \begin{pmatrix} \mathbf{g}_1(\tilde{\mathbf{y}}(t),\mathbf{s}(t),t|\phi^{g}) \\
    \mathbf{0} \\ \vdots \\ \mathbf{0} \\ \mathbf{g}_a(\tilde{\mathbf{y}}(t),\mathbf{s}(t),t|\phi^{g})\end{pmatrix} \rm{d}\mathbf{W}
\end{equation}
where the dimensionality of the function $\mathbf{f}(\cdot|\phi^{f}):\mathcal{R}^{\rm{N_x}(\rm{N}_{delay}+1)+\rm{N}_s}\rightarrow \mathcal{R}^{\rm{N_x}(\rm{N}_{delay}+1)}$ is analogous to the neural-ODE settings, while the non-zero elements of the function $\mathbf{g}(\cdot|\phi^{g})$ constitute a map $\mathcal{R}^{\rm{N_x}(\rm{N}_{delay}+1)+\rm{N}_s}\rightarrow \mathcal{R}^{(\rm{N_x}+\rm{N}_a)\times \rm{N}_W}$; $\mathbf{A}$ is a $\rm{N}_x \times \rm{N}_{W}$ dimensional matrix linking the auxiliary variables to the last state of the system. We notice how we are using a single network $\mathcal{R}^{\rm{N_x}(\rm{N}_{delay}+1)+\rm{N}_s}\rightarrow \mathcal{R}^{(\rm{N_x}+\rm{N}_a)\times \rm{N}_W}$ to parametrise the two terms $\mathbf{g}_1$ and $\mathbf{g}_a$ as depicted in the schemes of the neural-SDE architecture reported.  
The training of the neural-SDE follows a generative adversarial network paradigm, akin to previous works \cite{kidger2021neural}. For a detailed, didactic description of both Neural-ODE and neural-SDE models, we refer to the supplementary information.

\subsection*{Neural-SDEs as network nodes}
The augmented state $\mathbf{y}(t)$ is defined to describe the device as a Markovian system, and consequently to capture the system behaviour in a neural-ODE/SDE framework. As a consequence, $\mathbf{y}$ is a modelling abstraction, and does not correspond to the information exchanged between devices. 
Considering the point of view of the i-th device in the network, we can assume device interactions to occur via the variable $\mathbf{y}_i^{\pi}(t)=\boldsymbol{\pi}\big(\mathbf{y}_i(t) \big)$. The function $\boldsymbol{\pi}$ distinguishes the devices dynamics, $\mathbf{y}_i(t)$, from the quantities that dictate the exchange of information, $\mathbf{y}_i^{\pi}(t)$. We further notice how $\mathbf{\pi}$, the map between the abstracted $\mathbf{y}_i(t)$ and the physical quantity $\mathbf{y}^{\pi}_i(t)$, needs to be differentiable to permit backpropagation through the network dynamics. For a network of interacting devices and a specific task, the dynamics of node i can be described as:
\begin{align}
\rm{d}\mathbf{y}_i(t) &= \mathbf{f}_i\big(\mathbf{y}_i(t),\mathbf{s}_i(t),t\big)\rm{d}t + \mathbf{g}_i\big(\mathbf{y}_i(t),\mathbf{s}_i(t),t\big)\rm{d}\mathbf{W} \label{Eq.:Real_SDE_Main} \\
\mathbf{s}_i(t) &= \mathbf{h}_i\Big( \mathbf{y}^{\pi}_1(t),...,\mathbf{y}^{\pi}_{\rm{N}}(t),\mathbf{s}^{task}(t)|\boldsymbol{\theta}_i \Big) \label{Eq.:Interactions_Main}
\end{align}
where $\boldsymbol{\theta}_i$ is the subset of the network connectivity $\boldsymbol{\theta}$ that provides input to the i-th device and $\mathbf{s}^{task}(t)$ is a task-dependent driving stimulus. 
The sets of Eq.\ref{Eq.:Real_SDE_Main} and \ref{Eq.:Interactions_Main} can be used to compute the overall system activity across time assuming knowledge of all the $\mathbf{f}_i$ and $\mathbf{g}_i$. However, considering that knowledge of those functions is lacking in general settings, we approximate them through the trained models, simulating the following equations in place of Eq.\ref{Eq.:Real_SDE_Main} and Eq.\ref{Eq.:Interactions_Main} 

\begin{align}
\rm{d}\mathbf{y}_i(t) &= \mathbf{f}\big(\tilde{\mathbf{y}}_i(t),\tilde{\mathbf{s}}_i(t),t|\phi_i^{f}\big)\rm{d}t + \mathbf{g}\big(\tilde{\mathbf{y}}_i(t),\tilde{\mathbf{s}}_i(t),t|\phi_i^{g}\big)\rm{d}\mathbf{W} \label{Eq.:Fake_SDE_Main} \\
\tilde{\mathbf{s}}_i(t) &= \mathbf{h}_i\Big( \tilde{\mathbf{y}}^{\pi}_1(t),...,\tilde{\mathbf{y}}^{\pi}_{\rm{N}}(t),\mathbf{s}^{task}(t)|\boldsymbol{\theta}_i \Big) \label{Eq.:Fake_Interactions_Main}
\end{align}
where the device-dependent parameters $\phi_i^{f}$ 
and $\phi^{g}_i$ are used in place of the device dependent functions $\mathbf{f}_i$ and $\mathbf{g}_i$, and a tilde (\,$\tilde{.}$\,) is adopted to distinguish the experimental from the simulated quantities as before.
The connectivity parameters $\boldsymbol{\theta}$ are trained for a specific task via the gradients estimated from backward differentiation of Eq.\ref{Eq.:Fake_SDE_Main} and \ref{Eq.:Fake_Interactions_Main}. Performing this optimisation process assumes generated system responses $\tilde{\mathbf{y}}(t)$ are approximately equivalent to physical device activities $\mathbf{y}(t)$, and that the unknown devices dependencies $\dfrac{\rm{d}\mathbf{y}_i(t)}{\rm{d}\mathbf{s}_i(t')}$ can be approximated through 
$\dfrac{\rm{d}\tilde{\mathbf{y}}_i(t)}{\rm{d}\tilde{\mathbf{s}}_i(t')}$  $\forall t$ and $t'<t$ in the temporal interval considered. 
Backpropagation-through-time (BPTT), or truncated BPTT, will then decompose such total derivatives into the terms $\dfrac{\partial \tilde{\mathbf{y}}_i(t')}{\partial \tilde{\mathbf{s}}_i(t'-\delta t)}$ and $\dfrac{\partial \tilde{\mathbf{y}}_i(t')}{\partial \tilde{\mathbf{y}}_i(t'-\delta t)}$ $\forall t'$ in the considered interval. Optimisation is consequently executed on the simulated network, and the task-dependent optimal parameters are extracted like-for-like for use in experiments, where the resulting connectivity is validated on physically defined devices. The supplementary information provides more details on the use of BPTT and truncated-BPTT for the simulated system.

\subsection*{Nanomagnetic Ring Arrays (NRA)}
\subsubsection*{Fabrication of Ring Arrays}
Wafers of Si (001) with a thermally oxidised surface were spin-coated with 200nm of positive resist, with the nanoring array geometries and electrical contacts patterned via electron-beam lithography using a RAITH Voyager system. The magnetic nanoring arrays were patterned, then metallised to nominal thicknesses of 10nm via thermal evaporation of $\mathrm{Ni_{80}Fe_{20}}$ powder using a custom-built (Wordentec Ltd) evaporator (typical base pressures of below $10^{-7}$ mBar), before removal of the initial resist. Electrical contacts were patterned via a second lithography stage and were metallised via two-stage thermal evaporation of a 20nm Ti seed layer followed by a 100nm layer of Au. 
\subsubsection*{Electrical Transport Measurements of Ring Arrays}
Rotating magnetic fields were generated at 64 Hz via two pairs of air-coil electromagnets each with a voltage-controlled Kepco BOP 36-6D power supply. A sinusoidal voltage wave of 13,523 Hz was generated via an Aim-TTI instruments TG1000 signal generator and an SRS C5580 current source to generate 2 mA current, which was then injected to the nanoring arrays via the electrical contact pads. A National Instruments NI DAQ card measured the resulting potential difference across the device (modulated via anisotropic magnetoresistance (AMR) effects), sampling at 2 MHz.  Lock-in amplification was performed digitally by multiplying the measured voltage signal with a digitally generated reference wave matching the input current frequency, before filtering via a digital low-pass filter with a cut-off frequency of 320 Hz to remove the kHz component and leave the AMR dependent signal. The filtered waveform was then downsampled to a rate of 3.2kHz (50 samples per rotation of applied field) to reduce data size. 
\subsubsection*{Neural-SDE models of NRAs}
Training data for the deterministic component of the neural-SDE (parameterised by $\mathbf{f}$) was generated by driving the NRAs under 20,000 randomly generated sequences of 20 inputs, with the applied rotating magnetic fields spanning the responsive range of the devices, and recording the resulting AMR signals. The external signal $s(t)$ represented the magnitude of the applied field at time $t$.
Five delays were used here to define the hidden state, $\mathbf{y}(t)=\Big( x(t), ..., x(t-5)\Big)$, where the $x(t)$ represented the measured AMR signal. The data used to generated the stochastic component (parameterised by $\mathbf{g}$) was generated by driving the system with 1000 randomly generated input sequences of length 20 for 100 repetitions to generate example distributions of the noisy measurements, with 10 auxilliary variables used to generate different timescales of noise.
\subsection*{Artificial Spin Vortex Ices (ASVIs)}
Part of the description of the experimental methodologies for the ASVIs is reproduced from earlier works of several of the authors\cite{gartside2022reconfigurable}.
\subsubsection*{Fabrication of Artificial Spin Vortex Ices}
Artificial spin reservoirs were fabricated via electron-beam lithography liftoff method on a Raith eLine system with PMMA resist. 25 nm Ni$_{81}$Fe$_{19}$ (permalloy) was thermally evaporated and capped with 5 nm Al$_2$O$_3$. The flip-chip FMR measurements require mm-scale nanostructure arrays. Each sample has dimensions of roughly $\sim$ 3x2 mm. As such, the distribution of nanofabrication imperfections termed `quenched disorder' is of greater magnitude here than typically observed in studies on smaller artificial spin systems, typically employing 10-100 micron-scale arrays. The chief consequence of this is that the Gaussian spread of coercive fields is over a few mT for each bar subset. Smaller artificial spin reservoir arrays have narrower coercive field distributions, with the only consequence being that optimal applied field ranges for reservoir computation input will be scaled across a corresponding narrower field range, not an issue for typical 0.1 mT or better field resolution of modern magnet systems. 
\subsubsection*{Spectral Fingerprinting of Artificial Spin-vortex Ices}
Ferromagnetic resonance spectra were measured using a NanOsc Instruments cryoFMR in a Quantum Design Physical Properties Measurement System. Broadband FMR measurements were carried out on large area samples $(\sim 3 \times 2~ \text{ mm}^2)$ mounted flip-chip style on a coplanar waveguide. The waveguide was connected to a microwave generator, coupling RF magnetic fields to the sample. The output from waveguide was rectified using an RF-diode detector. Measurements were done in fixed in-plane field while the RF frequency was swept in 10 MHz steps. The DC field was then modulated at 490 Hz with a 0.48 mT RMS field and the diode voltage response measured via lock-in. The experimental spectra show the derivative output of the microwave signal as a function of field and frequency \cite{gartside2022reconfigurable}.
\subsubsection*{Neural-SDE models of ASVIs}
Training data for the deterministic component of the neural-SDE was gathered for a sequence of 13,000 inputs, with saturation pulses provided sporadically to reset the device. Data for the stochastic component was generated via 100 repetitions of 100 different input sequences of length 20. The ASVI response $\mathbf{x}(t)$ corresponds to the measured FMR spectra driven by an external field of amplitude $s(t)$. The dimensionality of $\mathbf{x}$ was sufficient to capture the system's dynamics and augmentation of the N-DE input variables was not necessary, setting $\mathbf{y}(t)=\mathbf{y}^{\pi}(t)=\mathbf{x}(t)$. 
\subsection*{Physical Neural Networks}
\subsubsection*{Physical systems as dynamic nodes}
The NRA devices were initialised via a single rotation of magnetic field at 80 Oe, with a sample from the distribution of the final AMR states of the initialisation procedure used as initial conditions for the neural-SDE model. The ASVI devices were initialised via a linearly applied field of 235 Oe, with similar selection of initial conditions for the model. The feed-forward networks were constructed by repeated measurements of the physical device, mimicking the flow of information through the network. Input data were combined with the transferred weights, then encoded into the strength of the applied magnetic fields and provided to the devices node-by-node within a given layer. The outputs of each layer were then combined with their respective weights and passed to the next layer in the network where the process was repeated. This hybrid between digitally stored network weights and physical nodes is due to current experimental limitations rather than limitations of the framework. 
\subsubsection*{Partially Observable MNIST}
The data for this task converts the original 784 dimensional input of the MNIST digits into a sequence of length N with 784 input features per step. At each timestep, the information from 784/N pixels is given via random sampling, and removed from the sample pool for subsequent images in the sequence. Hence, all of the information from the digit is provided by the end of the sequence. There is no correlation between the sampling process across multiple digits, resembling different permutations of information for every digit. Classification occurs from activities at the end of the sequence only. Training was performed via backpropagation through time in simulation, with hyperparameters tuned against a small validation set also in simulation. Testing was performed on networks of real devices on 1000 samples of unseen data, with reported accuracies averaged over three experimental runs with different masking of data.
\subsubsection*{Movement Classification of a Neuroprosthetic Device}
For the neuroprosthetic task, we adopted the second classification task (exercise B) from the Ninapro database \cite{atzori2014electromyography}, where sEMG activities have been recorded for 27 subjects. In these settings, the physical neural network is asked to perform gesture recognition from the sEMG recordings for all subjects. 
The sEMG temporal data has been preprocessed through a Kalman filter and sub-sampled at 100 Hz, leading to input sequences of 30 time steps for each gesture. The data was split into training, validation and testing sets, where the validation set was used to tune the hyperparameters. For optimisation, we adopted truncated backpropagation through time to reduce simulation time and memory requirements. Particularly, we fixed the number of temporal steps over which dependencies are considered and BPTT is carried out to ten. The seventeen classes were represented via one-hot encoding of output neurons, with the target signal of the same length as the input data. Training was performed by optimising the model output over a reduced window of the original signal, corresponding to the most meaningful information on the gesture (timestep 12 to 18). This meant that parts of the signal which are not informative for classification do not disrupt the learning process. This produced a single model which was able to perform decently over many time steps of prediction,  shown in Figure \ref{fig:Supp_pNeuro} in the supplementary material. 

\subsubsection*{Mackey--Glass Future Prediction and Cascade Learning}
The signals used for prediction were generated via the following delay-differential equations \cite{stenning2024neuromorphic, glass2010mackey}:
$\frac{dx}{dt} = \frac{\alpha x(t-\tau)}{1+x(t-\tau)^{n}} - \beta x(t)$, with $\alpha = 0.2, \beta = 0.1, \tau = 17, n = 10$, and $x_0 = 1.2$, solved numerically with a fourth-order Runge-Kutta solver and a timestep $dt=2$, producing quasi-periodic behaviour on the order of 25 samples. 5100 samples were generated, with the first 100 samples discarded as a wash-out of initial conditions. The next 1000 samples were used for training, and the same 1000 samples were used to gather the corrective datasets for the cascade learning approach. Model performance was evaluated in experiments over the remaining 4000 samples in 10 subsections of 400 points, with the resulting error bars reflecting performance over the 10 sections.

\subsection*{Author Contributions}
L.M. and M.O.E. developed the Neural-SDE framework. L.M., I.T.V., and K.D.S. performed all model fitting to experimental devices. I.T.V. performed experimental measurements of nanorings. K.D.S. performed experimental measurements of spin-ices. C.S. and G.V. fabricated nanoring samples. K.D.S. and J.C.G. fabricated spin-ice samples. I.T.V., C.S., and G.V. developed the experimental setup for nanoring measurement. D.G. optimised Neural-SDE code. K.D.S, L.G., D.S., and D.D. developed preliminary LSTM models for spin-ices. D.H. performed preprocessing of neuroprosthetic data. L.M., I.T.V., M.O.E., and E.V. drafted the manuscript. K.D.S, C.S., G.V., S.S., and J.C.G. provided primary reviewing and editing of manuscript. T.J.H. and W.R.B. provided feedback on experimental measurements for nanorings and spin-ices respectively. E.V. provided technical input on model development throughout the project. All authors assisted in reviewing and editing of the manuscript. L.M. and E.V. conceived the work. 
\subsection*{Acknowledgements}
L.M., I.V., T.J.H., M.O.E and E.V. acknowledge funding from the EPSRC MARCH Project No. EP/V006339/1. M.O.E, T.J.H and  E.V. also acknowledge funding from the EPSRC Project No. EP/S009647/1. D.G. and S.S. acknowledge funding from the EPSRC MARCH Project No. EP/V006029/1. K.D.S. was supported by The Eric and Wendy Schmidt Fellowship Program and the Engineering and Physical Sciences Research Council
(Grant no. EP/W524335/1). G.V. and T.J.H. were funded via Horizon 2020 FET-Open SpinEngine (Agreement No 861618). J.C.G. was supported by the Royal Academy of Engineering under the Research Fellowship programme and the EPSRC ECR International Collaboration Grant ‘Three-Dimensional Multilayer nanomagnetic Arrays for Neuromorphic Low-Energy Magnonic Processing’ EP/Y003276/1.

\label{Bibliography}
\bibliography{bib.bib}
\clearpage
\section{Supplementary Information}\label{sec4}

\subsection{Results of NSDE Modelling}
\begin{figure*}[t!]
\includegraphics[width=0.95\columnwidth]{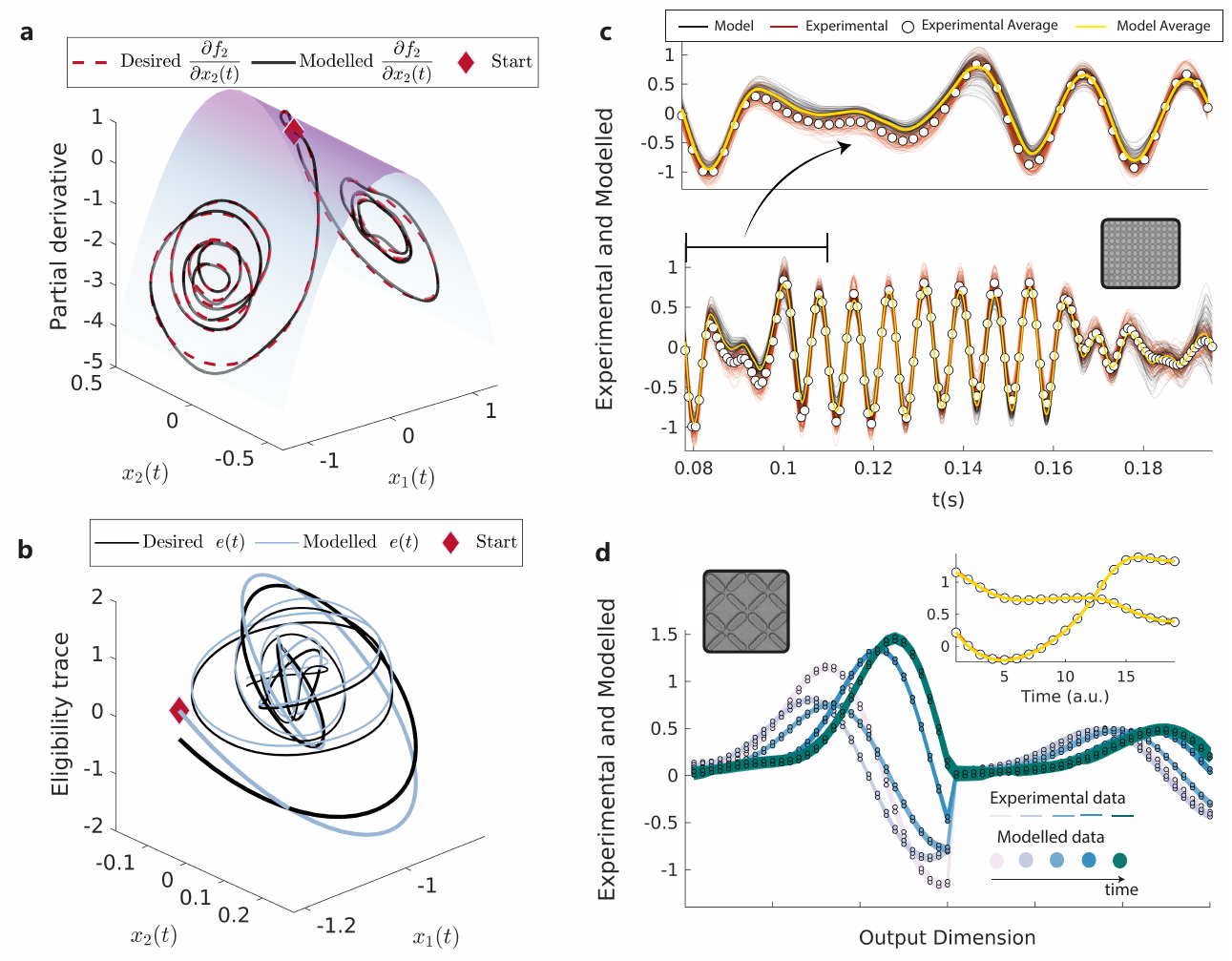}
\centering
\caption{\textbf{Modelling of simulated and experimental dynamical systems.} \small{
Panel (\textbf{a}) illustrates a simulated example of a partial derivative of the acceleration ($f_2$) with respect to position for the Duffing oscillator. The surface represents the partial derivative as position ($x_1(t)$) and velocity ($x_2(t)$) of the system vary. The example trajectories compare the gradient over time, calculated both analytically (red) and via differentiation of the Neural-SDE model (black), for two input sequences, showing excellent agreement.
Panel (\textbf{b}) provides a more general view of the model’s ability to act as a surrogate for device gradients; here, we adopt eligibility traces that accumulate gradient information (see Main text and Supplementary Information for more details). The difference between the desired and modelled eligibility traces increases due to error accumulation. Panel (\textbf{c}) compares responses generated via the Neural-SDE model (black and yellow lines) and experimentally gathered data of the NRA device (red lines, white circles) for 100 repetitions of a random input sequence. Panel (\textbf{d}) illustrates the Neural-SDE's ability to model the high-dimensional, experimentally measured responses of an artificial spin-vortex-ices (ASVI) device. Here, the x-axis corresponds to the different output dimensions of the device responses, while the colours reflect the temporal evolution. Even for this multivariate system, the model (coloured lines) accurately captures the system behaviour (dots).}  }      
\label{fig:Fig2}
\end{figure*}

Two dynamical systems with well-defined analytical descriptions---the nonlinear leaky integrator and the Duffing oscillator---were used as control cases to test the model fitting process shown in Figure \ref{fig:Fig2}. These systems provided `ground-truth’ data, allowing for precise control over timescales and noise magnitudes, and enabling direct calculation of gradients with respect to input and node dynamics. This capability is crucial for assesing the model's accuracy in approximating the gradients required for optimisation. In contrast, these properties cannot be explicitly evaluated in physical systems due to the absence of differentiable mathematical descriptions and presence of experimental noise in measurements of state. Detailed definitions of the simulated systems’ dynamics, along with figures demonstrating the neural-SDE model’s high accuracy in capturing their behaviors, are available in the supplementary information.\\\\
To assess the model’s ability to approximate the terms needed for backpropagation through time, Figure \ref{fig:Fig2}(a) illustrates the partial derivative of the acceleration with respect to position for the Duffing oscillator, shown as a function of position and velocity. The surface depicts the parabolic trend of this relationship, while the lines represent temporal trajectories for two different external signals. The simulated trajectories from the analytical model (red dashed lines) closely match those derived from the Neural-SDE model’s backward pass (black lines).
\\\\
\subsection*{Challenges in computing with physical dynamics}
From the standpoint of each device within the network, the optimisation of the external signal necessitates a consideration of its dynamic nature. Achieving a mathematically precise optimisation process entails unravelling the system dependencies backwards in time, a procedure known as backpropagation through time (BPTT \cite{werbos1990backpropagation}). It is worth noting that any algorithm, including recent approximations of BPTT\cite{bellec2020solution,bellec2019biologically}, designed to execute such optimisation in the context of neural networks, assumes a certain mathematical understanding of how a system at time $t$ depends on its past states. In mathematical terms, defining the i-th system dynamics and the external signal at time $t$ through the variables $\mathbf{y}_{i}(t)$ and $\mathbf{s}_{i}(t)$ respectively, input optimisation necessitates knowledge of the total derivatives 
$\dfrac{{\rm d}\mathbf{y}_{i}(t)}{{\rm d} \mathbf{s}_{i}(t')}$ for any $t'<t$. BPTT and its variants untangle these total derivatives through a chain rule involving the partial derivatives $\dfrac{\partial \mathbf{y}_i(t)}{\partial \mathbf{y}_i (t-\delta t)}$ and $\dfrac{\partial \mathbf{y}_i(t)}{\partial \mathbf{s} (t)}$ for all $t$ in the considered temporal interval. Therefore, estimating these factors is crucial for training the network of interacting devices in a general setting.

Another challenging aspect in optimising interactions between physically defined dynamic systems lies in their intrinsic stochastic behaviours. Stochasticity can significantly impact the performance of networks, a challenge that becomes more apparent in dynamic systems where current noise realisations (or even stochasticity in the initial conditions) may abruptly influence future system behaviour. Consequently, we will adopt and formulate models that can also capture the stochasticity of the devices considered and provide an estimate of the uncertainty associated with a particular interaction.

\subsection{The formulated neural-SDE}
The framework outlined here leverages variants of neural ordinary/stochastic differential equations as differentiable models to capture the dynamics and stochasticity of devices. A Neural-ODE/SDE model was used to simulate a type of device, then a network of Neural-ODE/SDEs was adopted to simulate a network of interacting devices. The differentiation through the simulated system provided estimates of the necessary components to perform BPTT and to train the simulated interactions. The optimised parameters will be then transferred to the physically defined systems, where the performance is assessed. The following paragraph outlines the digital twins and introduces the required formalism.

Let us consider a dynamic system and the collection of information over experimentally measurable variables, denoted as $\mathbf{x}(t)$, representing the evolution of pertinent properties of the system under the influence of an external signal $\mathbf{s}(t)$.

In the initial phase, trajectories of responses $\Big(...,\mathbf{x}(t),\mathbf{x}(t+\delta t),...\Big)$ from the studied device are recorded for various input sequences $\Big(...,\mathbf{s}(t),\mathbf{s}(t+\delta t),...\Big)$. The range and statistics of these input sequences are deliberately defined to gather a comprehensive dataset of input/output responses. This dataset of trajectories is then employed to optimise the N-DE models, whose activities, which we denote as $\tilde{\mathbf{x}}(t)$, are trained to reproduce the corresponding real variables $\mathbf{x}(t)$.

Given the feed-forward structure of N-DEs in predicting the "next" system activity and the resulting Markovian prerequisites for their input representation, we enhance their input by incorporating delayed observations, as in prior works \cite{chen2022forecasting} on neural-ODEs. We define this augmented state at time $t$ as $\tilde{\mathbf{y}}(t)=\Big( \tilde{\mathbf{x}}(t), \tilde{\mathbf{x}}(t-\delta t),...,\tilde{\mathbf{x}}(t-\rm{N}_{\text{delay}}\delta t) \Big)$, and the corresponding real variable as $\mathbf{y}(t)=\Big( \mathbf{x}(t), \mathbf{x}(t-\delta t),...,\mathbf{x}(t-\rm{N}_{\text{delay}}\delta t) \Big)$.
While neural-ODEs can replicate deterministic dynamics \cite{chen2018neural,chen2022forecasting} and are traditionally trained with cost functions as mean-squared error, neural-SDE can additionally capture device stochastic behaviour \cite{kidger2021neural}.

\subsubsection*{Neural-ODE and SDE models}
The section is dedicated to providing more details on the procedure adopted in training Neural-ODE and SDEs.
Let us consider a dynamic system described by an $\rm{N_X}$-dimensional observable variable $\mathbf{x}(t)$ and driven by an $\rm{N_S}$-dimensional external signal $\mathbf{s}(t)$. Introducing the variable $\mathbf{y}(t)=\big(\mathbf{x}(t), \mathbf{x}(t-\delta t),..., \mathbf{x}(t-\rm N_{delay}\delta t)$, which corresponds to an augmented version of the dynamic of $\mathbf{x}(t)$ aimed to capture higher order moments and to provide a Markovian representation of the evolution of the system, we can define a dataset of transitions $\mathcal{D}=\Big \{ ...,\big(\mathbf{y}(t-\delta t), \mathbf{s}(t), \mathbf{y}(t) \big),... \Big\}$. The value of $\rm{N}_{delay}$ specifies the number of delayed activities concatenated in the vector $\mathbf{y}$, which consequently $\in \mathcal{R}^{\rm{N_X}\times(\rm{N}_{delay}+1)}$.

Considering the evolution of a variable $\mathbf{y}(t)$ across time, a neural-ODE corresponds to an ordinary differential equation parametrised via a neural network of the form
\begin{equation}\label{Eq.N-ODE}
    \begin{alignedat}{3}
    {\rm d}\tilde{\mathbf{y}}(t)=\mathbf{f}\big(\tilde{\mathbf{y}}(t),\mathbf{s}(t),t|\phi^{f}\big) {\rm d}t
    \end{alignedat}
\end{equation}
where $\phi^{f}$ are the parameters of the network, the tilde $(\, \tilde{.}\,)$ is introduced to distinguish the target from the generated variables, and $\mathbf{s}(t)$ is an external signal.
As such, given an initial condition $\mathbf{y}(t_0)$ and a driving signal $\mathbf{s}(t)$, the neural-ODE will generate trajectories $\tilde{\mathbf{y}}(t)$ through iterative numerical integration. The input to the neural network $\mathbf{f}$ is thus a vector containing $\tilde{\mathbf{y}}$, $\mathbf{s}$ and a measure of time (which can also be included in the signal $\mathbf{s}(t)$). Neglecting the latter temporal information, the dimensionality of the input to the $\mathbf{f}$ network is $\rm{N_X}\times(\rm{N}_{delay}+1)+\rm{N_S}$. For a better exposition, it is now useful to expand the above equation into its different delayed components
\begin{equation}
    \begin{pmatrix} {\rm d}\tilde{\mathbf{x}}(t) \\ {\rm d}\tilde{\mathbf{x}}(t-\delta t) \\ \vdots \end{pmatrix}=
    \begin{pmatrix} \mathbf{f}_1(\tilde{\mathbf{y}}(t),\mathbf{s}(t),t|\phi^{f}) \\
    \mathbf{f}_2(\tilde{\mathbf{y}}(t),\mathbf{s}(t),t|\phi^{f}) \\ \vdots\end{pmatrix}{\rm d}t
\end{equation}
where we used the term $\phi^{f}$ for the different parameters defining $\mathbf{f}_1,\mathbf{f}_2,...$ for simplicity of notation. To be more precise, the parameters $\phi^{f}$ are shared among the $\mathbf{f}_i$ functions but for the different read-outs, which lead to the various  $\mathbf{f}_i$ in the output layer of the network. This notation is also  adopted in other equations defining the neural-SDE models.
We observe how, given that optimisation of $\tilde{\mathbf{y}}(t)$ should lead such a variable to contain delayed system activities, it would be possible to impose a solution to $\mathbf{f}_2,...,\mathbf{f}_{\rm{N}_{delay}+1}$ a priori. An example of this would correspond to set $\mathbf{f}_2=\dfrac{\tilde{\mathbf{x}}(t)-\tilde{\mathbf{x}}(t-\delta t)}{\delta t}$, or in other words to exploit our knowledge that the evolution of $\tilde{\mathbf{x}}(t-\delta t)$ is $\tilde{\mathbf{x}}(t)$, which the model has already computed.
In practice, we find that such an imposition is not necessary and that allowing training of $\mathbf{f}_2,...,\mathbf{f}_{\rm{N}_{delay}+1}$ can marginally improve performance while the system will naturally discover to act as a shift register for the delayed variables. 

Considering now the above-defined dataset $\mathcal{D}$, we can sample segments of trajectories with random starting times $t_0$ to define a batch of sequences used as targets for the model. The stochasticity in the selection of the starting times permits the batch sequence dynamics to be uncorrelated, and it is particularly advised for long temporal dynamics. For each sequence, the neural-ODE is initialised at the values $\tilde{\mathbf{y}}(t_0)=\mathbf{y}(t_0)$ and will generate a response $\tilde{\mathbf{y}}(t)$ for the intervals considered. Optimisation of the parameters $\phi^{f}$ is achieved through the minimisation of a cost function that reflects the discrepancy between the target and the generated sequences. Defining $\delta_j(t)=\tilde{\mathbf{y}}_j(t)-\mathbf{y}_j(t)$, the mean-squared error function to be minimised for a specific sequence is
\begin{equation}
    \mathcal{L}_{\phi^{f}}=\sum_t \sum_j \delta_j(t)^2
\end{equation}
where the $t \in$ to the interval $[t_0 \ t_0+\rm{T}]$ considered, where $\rm{T}$ defines the temporal length of the segment of the sequence in question.
In practice, the error function will be defined and averaged across all segments of the considered minibatch to iteratively update $\phi^{f}$. 

While neural-ODE are effective at capturing deterministic dynamics, a neural-SDE aims to additionally model the stochastic behaviour of the system in consideration, and it is defined as
\begin{equation}\label{Eq.N-SDE}
    \begin{alignedat}{3}
    {\rm d}\tilde{\mathbf{y}}(t)=\mathbf{f}\big(\tilde{\mathbf{y}}(t),\mathbf{s}(t),t|\phi^{f}\big){\rm d}t
    +\mathbf{g}\big(\tilde{\mathbf{y}}(t),\mathbf{s}(t),t|\phi^{g}\big){\rm d}\mathbf{W}
    \end{alignedat}
\end{equation}
where $\rm{d}\mathbf{W}$ is the Wiener process, and $\phi^{g}$ are the parameters reletad to the stochastic component. This model is thus parametrised by two networks $\mathbf{f}$ and $\mathbf{g}$, whose inputs are the $\rm{N_X}\times(\rm{N}_{delay}+1)+\rm{N_S}$ dimensional concatenation of $\tilde{\mathbf{y}}$ and $\tilde{\mathbf{s}}$. In this case, we generally found that it is important to constrain the function $\mathbf{g}$ to generate stochasticity only on the more recent $\rm{N_X}$-dimensional vector $\mathbf{x}(t)$, mathematically leading to

\begin{equation} \label{Eq.N-SDE_delayed_0aux}
    \begin{pmatrix} 
    {\rm d}\tilde{\mathbf{x}}(t) \\ 
    {\rm d}\tilde{\mathbf{x}}(t-\delta t) \\ 
    \vdots \\ 
    {\rm d}\tilde{\mathbf{x}}(t-{\rm N_{delay}}\delta t) 
    \end{pmatrix}=
    \begin{pmatrix} 
    \mathbf{f}_1(\tilde{\mathbf{y}}(t),\mathbf{s}(t),t|\phi^{f}) \\
    \mathbf{f}_2(\tilde{\mathbf{y}}(t),\mathbf{s}(t),t|\phi^{f}) \\ \vdots \\ 
    \mathbf{f}_{\rm{N}_{delay}+1}(\tilde{\mathbf{y}}(t),\mathbf{s}(t),t|\phi^{f}) 
    \end{pmatrix}{\rm d}t
    +
     \begin{pmatrix} 
     \mathbf{g}_1(\tilde{\mathbf{y}}(t),\mathbf{s}(t),t|\phi^{g}) \\
    \mathbf{0} \\ \vdots \\ \mathbf{0}
    \end{pmatrix} {\rm d}\mathbf{W}
\end{equation}
The reason for this lies in the additive nature of the noise and can be understood by envisioning the flow of information in the system through the delays. As the predicted activity $\tilde{\mathbf{x}}(t+\delta t)$ at time $t$ shifts through the $\mathbf{f}_i$ functions to $\tilde{\mathbf{x}}(t-\delta t)$ at time $t+\delta t$, $\tilde{\mathbf{x}}(t-2 \delta t)$ at time $t+2\delta t$ and so on, the stochastic function should not additionally contribute to value of such a variable multiple times. 

 Having defined the basic neural-SDE model, we leave the details of the learning algorithm and of the use of auxiliary variables to the next section.

\begin{figure*}[ht!]
\includegraphics[width=0.95\columnwidth]{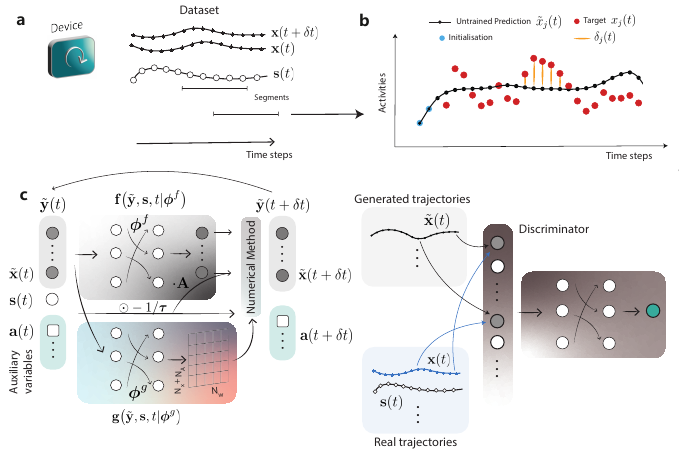}
\caption{\textbf{a} A scheme of the process of gathering a dataset used to fit an N-DE model. The trajectories are organised such that the concatenation of $\mathbf{s}(t_0)$ and $\mathbf{y}(t_0)$ are the initial conditions of the N-DE model, which is asked to produce trajectories that mimic the output $\mathbf{y}(t+\delta t)$ across time.
The process is depicted in panel \textbf{b} for an example segment of trajectories. The initial conditions are depicted in blue and correspond to the starting activities of the N-DE, whose generated response is reported in black. The optimisation process of a neural-ODE typically involves minimisation of the mean-squared error of $\delta (t)$, which is simply the difference between the generated and target activities, over the mini-batch of segments considered.  }
\label{fig:Fig1_Sup}
\end{figure*}

\subsubsection*{Neural-SDE optimisation and coloured noise}
We notice how in the above definition of the neural-SDE (Eq.\ref{Eq.N-SDE_delayed_0aux}) the stochastic realisations of $\mathbf{g}_1\mathbf{\rm{d}\mathbf{W}}$ are integrated through the temporal kernel defined by $\mathbf{f}_1$. This leads the model to exhibit a limited ensemble of autocorrelation structures and its inability to capture coloured noise. To circumvent this limitation, we augmented the neural-SDE representation through auxiliary variables $\mathbf{a}(t)$ that operate over a wide range of timescales. Augmentation of stochastic differential equations with Ornstein-Uhlenback process is a commonly adopted strategy to introduce richer noise characteristics.
The adapted neural-SDE equations, already expanded for the different delayed inputs, are
\begin{equation} \label{Eq.N-SDE_delayed_0_aux}
    \begin{pmatrix} {\rm d}\tilde{\mathbf{x}}(t) \\ 
    {\rm d}\tilde{\mathbf{x}}(t-\delta t) \\ 
    \vdots \\ 
    {\rm d}\tilde{\mathbf{x}}(t-{\rm N_{delay}}\delta t) \\ {\rm d}\mathbf{a}(t)
    \end{pmatrix}=
    \begin{pmatrix} \mathbf{f}_1(\tilde{\mathbf{y}}(t),\mathbf{s}(t),t|\phi^{f})+\mathbf{A}\mathbf{a}(t) \\
    \mathbf{f}_2(\tilde{\mathbf{y}}(t),\mathbf{s}(t),t|\phi^{f}) \\ \vdots \\ 
    \mathbf{f}_{{\rm N_{delay}}+1}(\tilde{\mathbf{y}}(t),\mathbf{s}(t),t|\phi^{f}) \\ -\dfrac{1}{\boldsymbol{\tau}}\mathbf{a}(t) \end{pmatrix}{\rm d}t+
     \begin{pmatrix} \mathbf{g}_1(\tilde{\mathbf{y}}(t),\mathbf{s}(t),t|\phi^{g}) \\
    \mathbf{0} \\ \vdots \\ \mathbf{0} \\ \mathbf{g}_a(\tilde{\mathbf{y}}(t),\mathbf{s}(t),t|\phi^{g})\end{pmatrix} {\rm d}\mathbf{W}
\end{equation}
where $\mathbf{a}$ are the $\rm{N}_a$-dimensional auxiliary variables, $\boldsymbol{\tau}$ is the $\rm{N}_a$-dimensional vector defining the different timescales, $\mathbf{A}$ is a trainable $\rm{N_X}\times\rm{N_a}$ connectivity matrix that links the auxiliary variables to the current activities $\mathbf{x}(t)$. As before, we use the same notation for the parameters $\phi^{g}$ of the different terms $\mathbf{g}_1$ and $\mathbf{g}_a$ for simplicity of notation. We  indicate as $\mathbf{g}$ the overall function (in square brackets) that is multiplied by the Wiener process. A scheme of the corresponding network architecture is given in Figure \ref{fig:Fig1_Sup} c, depicting the networks $\mathbf{f}(\tilde{\mathbf{y}}(t),\mathbf{s}(t),t|\phi^{f}):\mathbb{R}^{\rm{N_X}(\rm{N}_{delay}+1)}\rightarrow \mathbb{R}^{\rm{N_X}(\rm{N}_{delay}+1)}$, $\mathbf{g}(\tilde{\mathbf{y}}(t),\mathbf{s}(t),t|\phi^{g}):\mathbb{R}^{\rm{N_X}(\rm{N}_{delay}+1)}\rightarrow \mathbb{R}^{(\rm{N_X}+\rm{N_a})\times \rm{N_W}}$, where we excluded the zero elements in the dimensionality of its output, and how they interact. 

Optimisation of the neural-SDE is accomplished through a generative adversarial network paradigm where the neural-SDE is treated as a generator similarly to previous works \cite{kidger2021neural} and a discriminator provides the feedback for optimisation. In our settings, the discriminator is a feedforward network, in the figure an MLP, receiving trajectories of dynamics and the corresponding driving signals as inputs. Denoting with $\mathbf{s}_d$ the input to the discriminator network, $\mathbf{s}^{g}_d=\Big( \mathbf{s}(t_0),\tilde{\mathbf{x}}(t_0),\mathbf{s}(t_0+\delta t),\tilde{\mathbf{x}}(t_0+\delta t),...,\mathbf{s}(t_0+\rm{T}),\tilde{\mathbf{x}}(t_0+\rm{T}) \Big)$  for the generated activities and  $\mathbf{s}^{r}_d=\Big(\mathbf{s}(t_0),\mathbf{x}(t_0),\mathbf{s}(t_0+\delta t),\mathbf{x}(t_0+\delta t),...,\mathbf{s}(t_0+\rm{T}),\mathbf{x}(t_0+\rm{T}) \Big)$ for the ``real'' trajectories, where we recall that $\rm{T}$ is the temporal length of the trajectory considered. Thus, the discriminator is a network $\mathbf{d}:\mathbb{R}^{(\rm{N_X}+\rm{N_S})(T+1)}\rightarrow \mathbb{R}$, whose output reflects a confidence estimate that the input data has not been generated by the neural-SDE.
The optimisation is achieved via a Wasserstein loss function with gradient penalty \cite{gulrajani2017improved}, which we recall here for completeness 
\begin{equation}
    \underset{\mathbf{s}^{g}} {\mathbb{E}} \{ d(\mathbf{s}_d^{g}) \}- \underset{\mathbf{s}^{r}} {\mathbb{E}} \{ d(\mathbf{s}_d^{r}) \}+(||\underset{{\hat{\mathbf{s}}}}{\nabla} d(\hat{\mathbf{s}}_d)||_2-1)^2 
\end{equation}
where the $\hat{\mathbf{d}}$ corresponds to linearly interpolated data between $\mathbf{s}^{r}_d$ and $\mathbf{s}^{g}_d$ \cite{gulrajani2017improved}.
The cost function for the generator is typically 
\begin{equation}
     -\underset{\mathbf{s}^{g}} {\mathbb{E}} \{ d(\mathbf{s}_d^{g}) \}
\end{equation}
However, we noticed that such a formulation can lead the training process to undesirable local minima because of mode collapses, i.e. the tendency of the network to only capture specific statistical properties of the data, or oscillations in the performance. Denoting with $\mathbf{d}^{(i)}$ the activities of the i-th layer of the discriminator, we reformulated the minimisation process of the generator in a contrastive learning fashion with the new cost function 
\begin{equation}
    ||\langle \mathbf{d}^{(i)}(\mathbf{s}^r)\rangle-\langle \mathbf{d}^{(i)}(\mathbf{s}^g)\rangle||^2_2+||\sigma \big( \mathbf{d}^{(i)}(\mathbf{s}^r) \big)-\sigma \big( \mathbf{d}^{(i)}(\mathbf{s}^g) \big)||^2_2 
\end{equation}
where $\langle \cdot \rangle$ and $\sigma( \cdot )$ are the average and standard deviation computed over the minibatch considered. We observed that this formulation led to slower convergence time of the algorithm but to improved solutions.
Finally, while the sets of parameters of a neural-SDE have been previously trained simultaneously via the discriminator, we found it necessary to improve our control over the solution. Optimisation of GANs can indeed require intensive fine-tuning \cite{saxena2021generative} because of non-converging behaviours and mode-collapse difficulties. To partially circumvent these problems, we pre-trained the deterministic component of the neural-SDE, i.e. the ODE part, with a mean-squared error function. In such a way, the network is already capable of capturing the average behaviour of the dynamics when we start the optimisation through the discriminator. Of course, this can introduce a trade-off on when to stop the deterministic training for all those cases where noise can dramatically affect the average system evolution. For instance, if a dynamic system can exhibit bifurcations, it is challenging to optimise the deterministic component, which might interpolate among the different possible trajectories after bifurcation. However, we found that this apparent complexity can be circumvented by augmenting the input to the model including more delayed activities (increasing $\rm{N}_{delay}$) and that pretraining the neural-ODE would still be less tedious than fine-tuning the parameters for end to end optimisation of the neural-SDE through the discriminator.  It is finally important to notice that optimisation of the N-DE models has been achieved by backpropagating through the numerical method as in previous neural-SDE formulations\cite{kidger2021neural}. 

In the context of capturing device dynamics, we also advise adopting statistical metrics computed on an appropriately prepared validation set to control and stop the optimisation of the neural-SDE. 
In particular, we generated trajectories of responses of the considered device when subjected to repetitions of the same (approximately the same) external signal. We repeated then this process for multiple signals, and computed metrics evaluating the discrepancy between the average, variance, and autocovariance structure between the generated and physical device responses. We then chose the model across training that corresponds to the lowest error, computed simply by summing across the different metrics.
The result of this procedure permitted a high degree of control over the solution.

\begin{figure*}[h!]
\centering
\includegraphics[width=0.89\columnwidth]{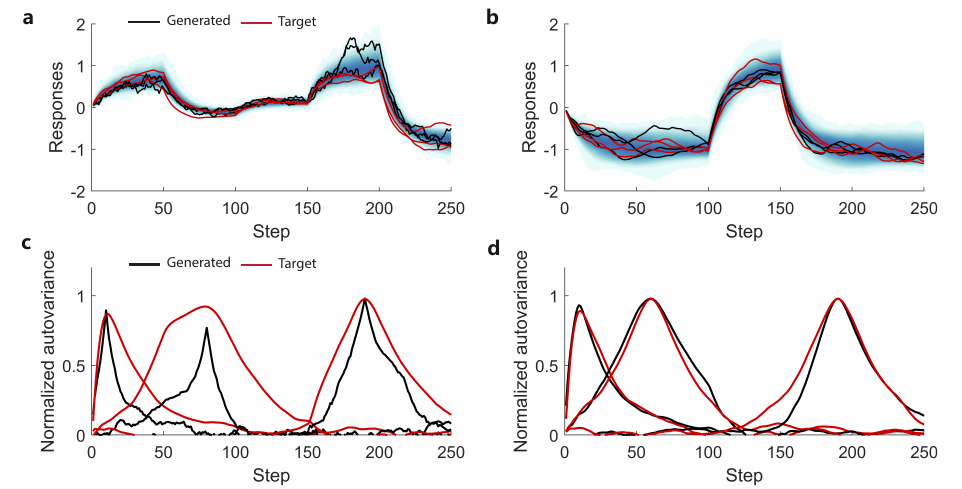}
\caption{The figure aims to illustrate the importance of capturing coloured noise for an improved expressivity of the neural-SDE. The dynamic system in question is a simple non-linear leaky integrator, where its stochasticity lies on a slower timescale in comparison to its leakage term, which defines its deterministic temporal kernel. Panels \textbf{a} and \textbf{c} (\textbf{b} and \textbf{d}) correspond to a neural-SDE without (with) augmentation of the auxiliary variables. \textbf{a} and \textbf{c} report examples of generated trajectories (black) compared to the corresponding system dynamics (red), where the blue area reflects the dispersion of the distribution of the dynamic process. Given that both models can capture the dispersion of the process, the dynamics generated by the neural-SDE of panel \textbf{a} exhibit rapidly changing stochastic behaviour, not capturing the smoother trends of the reference system. This translates into an inaccurate reproduction of the autocovariance structure, shown in panel \textbf{c}, where the different trends correspond to the diverse reference times over which the autocovariance function is computed. In contrast, the dynamics of the augmented neural-SDE are in considerably improved agreement with the process under study (panel \textbf{b} and \textbf{d}). Despite the simplicity of this example, the result shows how augmentation of the neural-SDE model can strongly improve its expressivity, and it might be necessary to capture stochastic behaviours that are not completely entangled to the temporal kernel of its deterministic component.}
\label{fig:Fig2_Sup}
\end{figure*}

\subsection{Optimisation of a dynamic physical neural network}

An optimised neural-SDE is used as a complex node of a network that simulates the ensemble of interacting devices.

\subsubsection*{Neural-SDE as nodes}

 The framework formulation assumes that the i-th device, considered in isolation, can be described as
\begin{equation}
{\rm d}\mathbf{y}_i(t)=
\mathbf{f}_i\big(\mathbf{y}_i(t),\mathbf{s}_i(t),t\big){\rm d}t
+\mathbf{g}_i\big(\mathbf{y}_i(t),\mathbf{s}_i(t),t\big){\rm d}\mathbf{W} \label{Eq.:Real_SDE}
\end{equation}
where the subscripts $i$ on the $\mathbf{f}$ and $\mathbf{g}$ functions are introduced to distinguish between the different devices' behaviours explicitly. As before, the augmented variable $\mathbf{y}_i$ is defined as $\mathbf{y}_i(t)=\Big(\mathbf{x}_i(t),\mathbf{x}_i(t-\delta t),...,\mathbf{x}_i(t-{\rm N_{delay}}\delta t)\Big)$, where $\mathbf{x}_i$ corresponds to the experimentally measurable time-dependent features. 

In the general context of a physically implemented network, the variables $\mathbf{y}_i(t)$ adopted to capture the devices' dynamics might not correspond to the information exchanged by the different devices.
We can instead safely assume that interactions occur via the variables  $\mathbf{y}_i^{\pi}=\boldsymbol{\pi} (\mathbf{y}_i(t))$, where $\boldsymbol{\pi}$ is an arbitrary projection leading to a lower dimensional space. 
The consideration of a network of interacting systems implies that 
\begin{equation} 
    \mathbf{s}_i(t)=\mathbf{h}_i\Big( \mathbf{y}^{\pi}_1(t),...,\mathbf{y}^{\pi}_{\rm{N}}(t),\mathbf{s}^{task}(t)|\boldsymbol{\theta}_i \Big) \label{Eq.:Interactions}
\end{equation}
where $\mathbf{s}^{task}(t)$ is a task-dependent input that drives the system, and $\mathbf{h}_i$ corresponds to the sub-network of nodes adjacent to the i-th device and that is parametrised by the connectivity $\boldsymbol{\theta}_i$. 
Depending on the specific network architecture, $\mathbf{h}_i$ might be a function of a subset of the above arguments. For instance, devices in the second layer of a typical feedforward structure will be connected only to the first layer of devices without receiving an external input $\mathbf{s}^{task}(t)$.
We denote as $\boldsymbol{\theta}$ the complete set of network parameters, while the system output is defined as a read-out $\mathbf{o}(t)=\mathbf{W}^{\rm{o}} \big( \mathbf{y}^{\pi}_1(t),..., \mathbf{y}^{\pi}_{\rm{N}}(t) \big)$. 
Given an example error function ${E}(t)=\big( \mathbf{o}(t)-\mathbf{o}^{\rm{task}}(t) \big)^2$, where $\mathbf{o^{\rm{task}}}$ is a task-dependent target, minimisation of ${E}(t)$ with respect to $\boldsymbol{\theta}$ leads to
\begin{equation}
    \dfrac{{\rm d}E(t)}{{\rm d}\boldsymbol{\theta}}=
    \dfrac{\partial E(t)}{\partial \mathbf{o}(t)} 
    \sum_i \dfrac{\partial \mathbf{o}(t)}{\partial \mathbf{y}^{\pi}_i(t)}
    \dfrac{{\rm d}\mathbf{y}^{\pi}_i(t)}{{\rm d} \boldsymbol{\theta}} \label{Eq.:BPTT}
\end{equation}
where $\dfrac{{\rm d}\mathbf{y}^{\pi}_i(t)}{{\rm d}\boldsymbol{\theta}}$ requires differentiation through the system's dynamics (Eq.\ref{Eq.:Real_SDE}) and of the terms $\mathbf{h}_i$ corresponding to the network structure. As a consequence, Eq.\ref{Eq.:BPTT} is intractable for physical systems that lack a precise differentiable description. 

A device $i$ belongs to a type of devices, whereas type is a set $\rm{D}$ of devices that have common statistical properties and whose behavioural differences are caused by fabrication variability ($\rm{D}$ might correspond to different ASVIs, for instance). The neural-SDE for device $i$ is
\begin{equation}
{\rm d}\tilde{\mathbf{y}}_i(t)=
\mathbf{f}_{d}\big(\tilde{\mathbf{y}}_i(t),\mathbf{s}_i(t),t|\phi^{f}_{d}\big){\rm d}t
+\mathbf{g}_{d}\big(\tilde{\mathbf{y}}_i(t),\mathbf{s}_i(t),t|\phi^{g}_{d}\big){\rm d}\mathbf{W} \label{Eq.:Approx_SDE}
\end{equation}
where the subscript $d$ indicates the presence of type-dependent functions and will be omitted for simplicity of notation. 
The external signal $\mathbf{s}_i(t)$ can be further augmented through a device identifier to account for fabrication variability, and we recall how the parameters of the neural-SDE are optimised to mimic the reference systems' behaviours as shown in the previous sections.  The optimised neural-SDEs are then adopted to estimate the unknown terms of Eq.\ref{Eq.:BPTT} through 
\begin{equation}
    \dfrac{{\rm d}E(t)}{\rm{d}\boldsymbol{\theta}} \approx \dfrac{\partial E(t)}{\partial \tilde{\mathbf{o}}(t)} \sum_i \dfrac{\partial \tilde{\mathbf{o}}(t)}{\partial \tilde{\mathbf{y}}^{\pi}_i(t)}
    \dfrac{{\rm d}\tilde{\mathbf{y}}^{\pi}_i(t)}{{\rm d} \boldsymbol{\theta}} \label{Eq.:BPTT_estimated}
\end{equation} 

While Eq.\ref{Eq.:BPTT_estimated} can be computed through auto-differentiation tools over the dynamics defined in Eq.\ref{Eq.:Approx_SDE}, it is still relevant to understand the mathematics behind the framework and expand the terms of Eq.\ref{Eq.:BPTT}. \newline

\subsubsection*{An isolated device and its eligibility trace}

\begin{figure*}[h!]
\centering
\includegraphics[width=0.8\columnwidth]{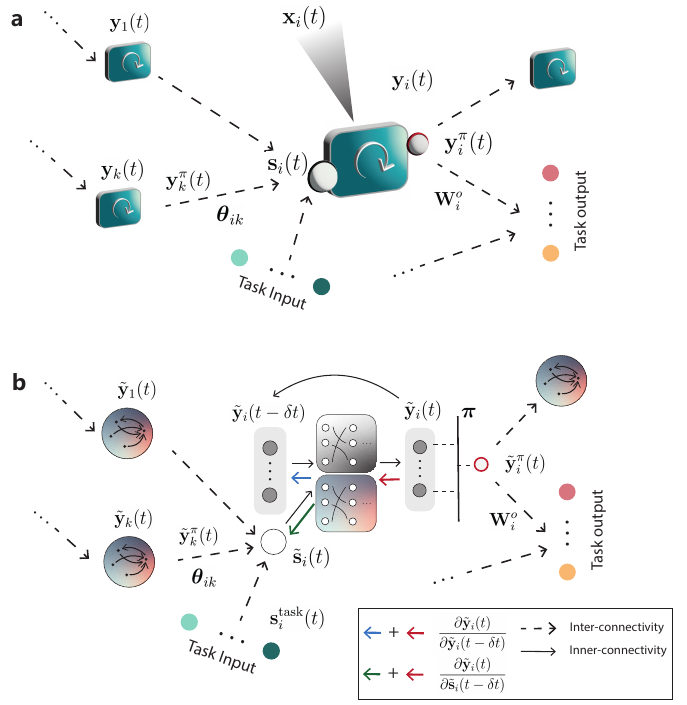}
\caption{Scheme of the network of devices (\textbf{a}) and of neural-SDEs (\textbf{b}) illustrating the formalism adopted (see text for more details). The central neural-SDE of panel \textbf{b} has been expanded to show inputs and outputs, highlighting the backward paths to estimate the dependencies (coloured arrows).   }
\label{fig:Fig3_Sup}
\end{figure*}

As a starting point and to introduce the eligibility traces of Figure \ref{fig:Fig2}(d) (Main text), consider an isolated device driven by a signal $\mathbf{s}^{task}(t)$ with activity $\mathbf{y}(t)$, whose total derivative can be unravelled in discrete time as
\begin{align}
     \dfrac{{\rm d}\mathbf{y}^{\pi}(t)}{\rm{d}\boldsymbol{\theta}}=\dfrac{\partial{\mathbf{y}^{\pi}(t)}}{ {\partial{\mathbf{y}(t)}}} \Bigg \{ \dfrac{ {\partial{\mathbf{y}(t)}}}{\partial{\boldsymbol{\theta}}}+\dfrac{\partial{\mathbf{y}(t)}}{\partial \mathbf{y}(t-\delta t)}\biggr \{\dfrac{\partial{\mathbf{y}(t-\delta t)}}{\partial \boldsymbol{\theta}}+\dfrac{\partial{\mathbf{y}(t-\delta t)}}{\partial \mathbf{y}(t-2\delta t)}\Bigl[ \dfrac{\partial{\mathbf{y}(t-2\delta t)}}{\partial \boldsymbol{\theta}}+...\Bigr]\biggr\} \Bigg \} \label{Eq.:Isolated_BPTT}
\end{align}
which can be estimated through the neural-SDE dynamics adopting the approximation $\mathbf{y}(t)=\tilde{\mathbf{y}}(t)$ in Eq.\ref{Eq.:Isolated_BPTT}.
Figure \ref{fig:Fig3_Sup} shows how $\dfrac{ {\partial{\tilde{\mathbf{y}}(t)}}}{\partial{\boldsymbol{\theta}}}$ and $\dfrac{\partial{\tilde{\mathbf{y}}(t)}}{\partial \tilde{\mathbf{y}}(t-\delta t)}$ are respectively computed differentiating the neural-SDE model, where the first term follows the backward path via the model's external input. If such terms are accurate estimators of the reference system regardless of the external input and previous activities, differentiation of the neural-SDE will be successful at optimising the connectivity $\boldsymbol{\theta}$.  A comparison of the Jacobians $\dfrac{\partial{\tilde{\mathbf{y}}(t)}}{\partial \tilde{\mathbf{y}}(t-\delta t)}$ and $\dfrac{\partial{\mathbf{y}}(t)}{\partial \mathbf{y}(t-\delta t)}$ is shown in the Main text, Figure \ref{fig:Fig2}(c), for the Duffing oscillator. 

Another more general approach to control the effectiveness of the model at capturing backward dependencies is to focus directly on the estimation of the total derivative $\dfrac{{\rm d}\mathbf{y}^{\pi}_i(t)}{{\rm d} \boldsymbol{\theta}}$. We can rewrite the terms of Eq.\ref{Eq.:Isolated_BPTT} as
\begin{equation}
    \dfrac{\partial{\mathbf{y}(t')}}{\partial{\boldsymbol{\theta}}}=\dfrac{\partial{\mathbf{y}(t')}}{\partial \mathbf{h}\big(\mathbf{s}^{task}(t')|\boldsymbol{\theta}\big)}\dfrac{\partial \mathbf{h}\big(\mathbf{s}^{task}(t')|\boldsymbol{\theta}\big)}{\partial \boldsymbol{\theta}}=\dfrac{\partial{\mathbf{y}(t')}}{\partial \mathbf{s}(t')}\mathbf{s}(t')\boldsymbol{\theta}^{-1}
\end{equation}
in which we assume $\mathbf{s}(t')=\mathbf{s}^{task}(t')\boldsymbol{\theta}$. This implies that Eq.\ref{Eq.:Isolated_BPTT} can be rewritten as 
\begin{equation}
    \dfrac{{\rm d}\mathbf{y}^{\pi}(t)}{{\rm d}\boldsymbol{\theta}}=
    \dfrac{\partial{\mathbf{y}^{\pi}(t)}}{ {\partial{\mathbf{y}(t)}}}\bigg \{ \dfrac{\partial{\mathbf{y}(t)}}{\partial \mathbf{s}(t)}\mathbf{s}(t)+\dfrac{\partial{\mathbf{y}(t)}}{\partial \mathbf{y}(t-\delta t)}\Big [\dfrac{\partial{\mathbf{y}(t-\delta t)}}{\partial \mathbf{s}(t-\delta t)}\mathbf{s}(t-\delta t)+...\Big] \bigg \} \boldsymbol{\theta}^{-1}=\dfrac{\partial{\mathbf{y}^{\pi}(t)}}{ {\partial{\mathbf{y}(t)}}} \mathbf{e}(t)\boldsymbol{\theta}^{-1}
\end{equation}

The variable $\mathbf{e}(t)$ is the eligibility trace shown in the Main text and can be computed \cite{bellec2020solution} through the recursive form $\mathbf{e}(t)=\dfrac{\partial \mathbf{y}(t)}{\mathbf{y}(t-\delta t)}\mathbf{e}(t-\delta t)+\dfrac{\partial \mathbf{y}(t)}{\partial \mathbf{s}(t)}\mathbf{s}(t)$, starting from $\mathbf{e}(t_0)=0$. In the main text, we compare $\mathbf{e}(t)$ with $\tilde{\mathbf{e}}(t)$, which is estimated through the neural-SDE of dynamic $\tilde{\mathbf{y}(t)}$.

\subsubsection*{Interacting devices and backpropagation}

We now expand Eq.\ref{Eq.:BPTT} considering the whole network of interacting devices.
Starting from the i-th node, BPTT through the network involves the chain of derivatives 
\begin{equation}
    \dfrac{{\rm d}\mathbf{y}_i(t)}{{\rm d}\boldsymbol{\theta}}=
    \dfrac{\partial{\mathbf{y}_i(t)}}{\partial{\boldsymbol{\theta}}}+\sum_k \dfrac{\partial{\mathbf{y}_i(t)}}{\partial \mathbf{y}_k(t-\delta t)}\biggr \{\dfrac{\partial{\mathbf{y}_k(t-\delta t)}}{\partial \boldsymbol{\theta}}+\sum_j \dfrac{\partial{\mathbf{y}_k(t-\delta t)}}{\partial \mathbf{y}_j(t-2\delta t)}\Bigl[ \dfrac{\partial{\mathbf{y}_j(t-2\delta t)}}{\partial \boldsymbol{\theta}}+...\Bigr]\biggr\}
\end{equation}
where, for instance, the $k$ elements of $\sum_k \dfrac{\partial{\mathbf{y}_i(t)}}{\partial \mathbf{y}_k(t-\delta t)} $ involve all the nodes connected to the $i$-th system, including the latter, and 
\begin{equation}
    \dfrac{\partial{\mathbf{y}_i(t)}}{\partial \mathbf{y}_k(t-\delta t)}=
    \dfrac{\partial{\mathbf{y}_i(t)}}{\partial \mathbf{h}_i\big(...,\mathbf{y}^{\pi}_k(t),...\big)}\dfrac{\partial \mathbf{h}_i\big(...,\mathbf{y}^{\pi}_k(t),...\big)}{\partial \mathbf{y}^{\pi}_k(t)}\dfrac{\partial \mathbf{y}^{\pi}_k(t)}{\partial \mathbf{y}_k(t)}\dfrac{\partial \mathbf{y}_k(t)}{\partial \mathbf{y}_k(t-\delta t)}, \ \forall k\neq i
\end{equation}


\subsubsection*{Initial conditions of physical devices}

As the initialisation of the activities of a dynamic system is a fundamental aspect of determining its dynamics, the considered devices need to be set to some starting initial conditions. This is particularly relevant in episodic settings, where each input sequence may require independent classification or correspond to distinct desired behaviours. 
Here, the devices are initialised to certain conditions at the beginning of each sequence. This is analogous to the starting hidden activities, typically set to zero, of a recurrent neural network within software models. While starting activities can be arbitrarily defined in software, experimental procedures naturally lead to a distribution of initial conditions. To accurately simulate the network of devices in software, it becomes relevant to experimentally obtain a probability distribution $\rm{p}_{IC}$ over initial conditions $\mathbf{y}_{\rm{IC}}=\Big( \mathbf{x}(t_0),...,\mathbf{x}(t_0+\rm{N}_{delay} \delta t) \Big)$ Subsequently, sampling is performed to establish the starting activity of each simulated device, replicating the process that would occur in experimental settings. 

Even in continuous settings, where the task can not be subdivided into independent sequences, the initial conditions still need to be set at least once. Considering devices that can be used for computational purposes and are driven by some signals, we might expect the relevance of the initial conditions to fade out at some point in time, similar to the Echo-state property \cite{jaeger_echo_2001, yildiz2012re}. However, we notice that this is not true in general for dynamic systems and that the process of appropriately setting the devices' starting activities would still be important in the temporal proximity of the initial conditions.

\subsection{Analytical Systems definitions}

The analytical systems adopted in this work are a nonlinear leaky integrator and a Duffing oscillator, both augmented with colored noise to study the expressive ability of the neural-SDE proposed and the effectiveness of the framework in dealing with complex stochastic characteristics.

\subsubsection*{Leaky Integrator}
The equation used for the leaky integrator are
\begin{align}
    {\rm d}x_1(t) & = \left[-\alpha_1 x_1(t)+\tanh(s(t))+x_2(t)+x_3(t) \right]{\rm d}t+
    \sigma_1 {\rm d W_1}  \label{Eq.:Leaky} \\  
    {\rm d}x_2(t) & = -\alpha_2 x_2(t){\rm d}t+\sigma_2 \tanh \left(x_1(t)\right){\rm d W_2} \nonumber \\ 
    {\rm d}x_3(t) & = -\alpha_3 x_3(t){\rm d}t+\sigma_3 \tanh\left(s(t)\right){\rm d W_3} \nonumber
\end{align}
where $\tanh(.)$ is the hyperbolic tangent, $\sigma_i$ are three constants reporting the variance of the different noise sources $\rm{d}W_i$, 
$\alpha_i$ the leakage terms, and $s(t)$ the external signal. While the first equation 
defines a typical leaky integrator with a non-linear activation function, the remaining terms introduce noise over different timescales and stochastic terms that are non-linear with respect to $x_1$ and $s$. For the majority of the simulations and if not stated otherwise, we set the discretisation step $\delta t=0.1$, and the values $\alpha_1=1$, $\alpha_2=0.5$, $\alpha_3=1.5$, $\sigma_2=\sigma_3=0.5$ and $\sigma_1=0.01$ and used stochastic Runge--Kutta 2 as integration method. The deterministic variant of the system is characterised exclusively by $x_1(t)$ without noise sources. 

To define the datasets used to train the neural-SDE model for this system, we need to define a statistically rich ensemble of external sequences. The driving signals adopted are square functions, whose values are randomly and uniformly sampled. Mathematically, $s(t)=s_i$ for $t \in [iT_{s} \ \ (i+1)T_{s}]$ $\forall i$, where $T_s$ defines the temporal length over which the external signal remains constant, and the values of $s_i$ are uniformly sampled in the interval $[-3 \ \ 3]$, whose range permits to sufficiently cover the regime where the hyperbolic tangent of Eq.\ref{Eq.:Leaky} varies. The training dataset for the neural-SDE comprises $1000$ sequences of temporal duration $200 \delta t$, where the length $T_s$ of the square waves is either $5$ or $20$. The adoption of different $T_s$ values permits adequate observation of the temporal kernel of the system and leads to an improved neural-SDE model, but it is not strictly necessary for the results reported. In this case, we assumed observation of the variable $x_1(t)$, and we defined the neural-SDE over the variable $y(t)=x_1(t)$ without augmentation of delayed observation of $x_1$ (see the N-DE definition for more details).    
Figure \ref{fig:Fig_Sup_analytical} (left panels) shows examples of trajectories generated by these systems and the responses generated by the optimised neural-SDE.

\subsubsection*{Duffing Oscillator}

The set of equations adopted for the Duffing oscillator are 
\begin{align*}
    {\rm d}x_1(t) & =x_2(t) {\rm d}t \\
    {\rm d}x_2(t) & =\big(-c x_2(t)+\alpha_1 x_1(t)+b x_1^3(t)+s(t)\cos(\omega t)\big){\rm d}t \\
    {\rm d}x_3(t) & = -\alpha_2 x_2(t){\rm d}t+\sigma_2 \tanh(x_1(t)){\rm d W_2} \\
    {\rm d}x_4(t) & = -\alpha_3 x_3(t){\rm d}t+\sigma_3 \tanh(s(t)){\rm d W_3}
\end{align*}

where the external signal is convolved through a sinusoidal function and takes the place of the parameter $\gamma$, used as a multiplicative factor of the sinusoidal function defining the Duffing oscillator. 
The stochasticity in the system has been introduced analogously to the non-linear leaky integrator defined above. We set $\alpha_1=1$, $\alpha_2=0.5$, $\alpha_3=1.5$, $\sigma_2=\sigma_3=0.05$ and $\sigma_1=0.01$ and adopted a discretisation step $\delta t=0.005$ and stochastic Runge--Kutta 4 as integration method. \newline 
Analogously to the leaky integrator, the driving external signal comprises square waves, where specific values of $s(t)$ are repeated for multiple time steps, i.e. $s(t)=s_i$ for $t \in [iT_{s} \ \ (i+1)T_{s}]$ $\forall i$. This allows the system to exhibit the characteristic behaviours of the Duffing oscillator while subjected to a temporally varying external signal. Thus, the setting implies that the operational timescale of the system is faster than the driving stimulus. The chosen values of $T_s$ defining the datasets of sequences are $20$ and $50$, while $s_i \in [-0.2 \ \ 0.2]$, over which the modelled Duffing oscillator can bifurcate in proximity of $s_i=\{-0.1, 0, 0.1$\}. As a consequence, the system will exhibit bifurcations that might be triggered by stochasticity.

\begin{figure*}[t!]
\centering
\includegraphics[width=0.85\columnwidth]{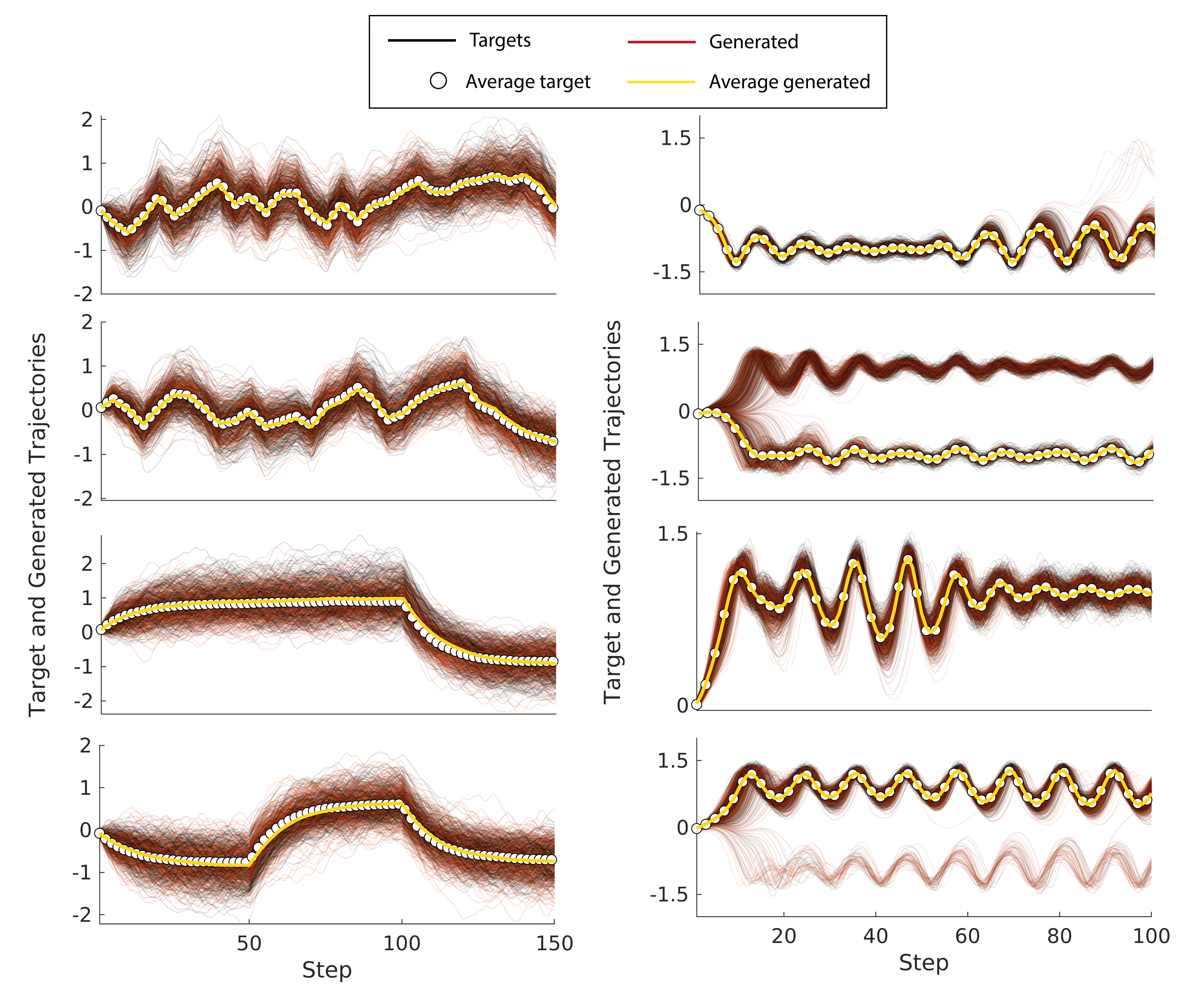}
\caption{Comparison between generated and target dynamics for the analytical systems considered, in particular for the leaky integrator (left) and Duffing oscillator (right). In each panel, the systems have been driven by an external signal $\mathbf{s}(t)$, while different panels correspond to different external signals. The trajectories generated by the neural-SDE are in black, while the reference trajectories are depicted in red. In each panel, the average response is also highlighted for comparison between the model (yellow line) and the analytical system (circles), where the average is conditioned to a specific bifurcating dynamic for the Duffing oscillator. Despite the variability in the deterministic and stochastic behaviours of the systems for different external signals and the presence of bifurcations, the neural-SDE is in striking agreement with the data.  }
\label{fig:Fig_Sup_analytical}
\end{figure*}

We assume observation of the system position $x_1(t)$ and velocities $x_2(t)$, but not of the other stochastic variables.
The input representation of the neural-SDE is $\mathbf{y}(t)=\Big ( x_1(t), x_2(t),  x_1(t-\delta t), x_2(t-\delta t), x_1(t-2\delta t), x_2(t-2\delta t) \Big)$, where we notice $\rm{N}_{delay}=2$. The training dataset comprises $2000$ sequences of temporal length $200 \delta t$.
Figure \ref{fig:Fig_Sup_analytical} compares trajectories sampled from the distribution of the stochastic variants of the leaky integrator (left panels) and  Duffing oscillator (right) with the trajectories generated by the neural-SDE. In each panel, the systems have been subjected to 100 repetitions of identical input sequences. The average of the neural-SDE generated (yellow lines) and reference trajectories (white circles) calculated analytically are in excellent agreement. Additionally, the distributions of dynamics generated by the model and analytically are equivalent through visual inspection, showing the ability of the model to replicate both deterministic and stochastic behaviour of the target system. The neural-SDE model can consequently capture various statistical properties of the target system, despite partial observation of the variables defining the systems in question. The stochastic Duffing oscillator can exhibit bifurcations at various instants with low probability ($<1\%$, bottom right panel). Crucially, the neural-SDE can approximately reproduce the rare event, the dynamics after the bifurcation, and its associated probability. To accurately capture this behaviour, the inclusion of delayed observations of the system activities proved to be necessary.

\subsection{Baseline Performance}

The section includes additional results for the different systems and tasks considered.

\subsubsection*{Partially observable MNIST task}
\begin{figure*}[h!]
\centering
\includegraphics[width=0.85\columnwidth]{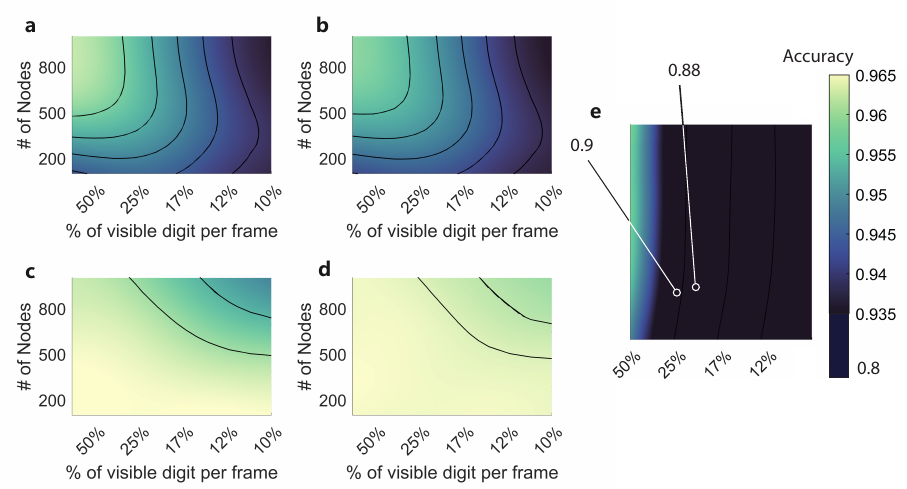}
\caption{Performance surface of a single-layer network (\textbf{a} and \textbf{b}) and a two-layer network (\textbf{c} and \textbf{d}) for interacting analytical leaky integrators. The reference networks are illustrated in panels \textbf{b} and \textbf{d}, while the twin networks are in \textbf{a} and \textbf{c}. Panel \textbf{e} reports the performance of a multilayer perceptron showing the quick decay in accuracy when the system does not have memory sources. }
\label{fig:Supp_perfGrids}
\end{figure*}

Here, we report the baseline performance for a network composed by the deterministic variants of the leaky integrator and Duffing oscillator, where the network connectivity is optimised through neural-ODEs.
The task faced is the partially observable MNIST of Figure \ref{fig:Fig3} (Main text).
Figure \ref{fig:Supp_perfGrids} (panels a,b,c,d) reports surfaces of performance for the leaky-integrator as the size of the network varies and for different percentages of observable digits (see Main text). Moreover, Figure \ref{fig:Supp_perfGrids}  shows the accuracy predicted by the twin networks (panels a and c) and exhibited by the reference networks (panels b and d), where the optimised connectivity is transferred. Considering the absence of stochasticity in the system and the simple behaviour of the leaky integrator, the simulated network is in striking agreement with the reference network. We can instead observe a trade-off between memory and non-linearity that is overcome by a network architecture that is composed of at least two hidden layers. The accuracy of the one-hidden layer architecture remains relatively low regardless of the network size. This trend is caused by the fact that, given the memory requirements of the task, a single layer would need to remember and non-linearly combine the successions of external inputs at the same time. Instead, the two-layer structure can surpass this limitation by adopting the first layer as a memory source and the second layer for non-linearity.  This is reflected by the high accuracy achieved by the two-layer structure over different network sizes.  

\begin{figure*}[h!]
\centering
\includegraphics[width=1.\columnwidth]{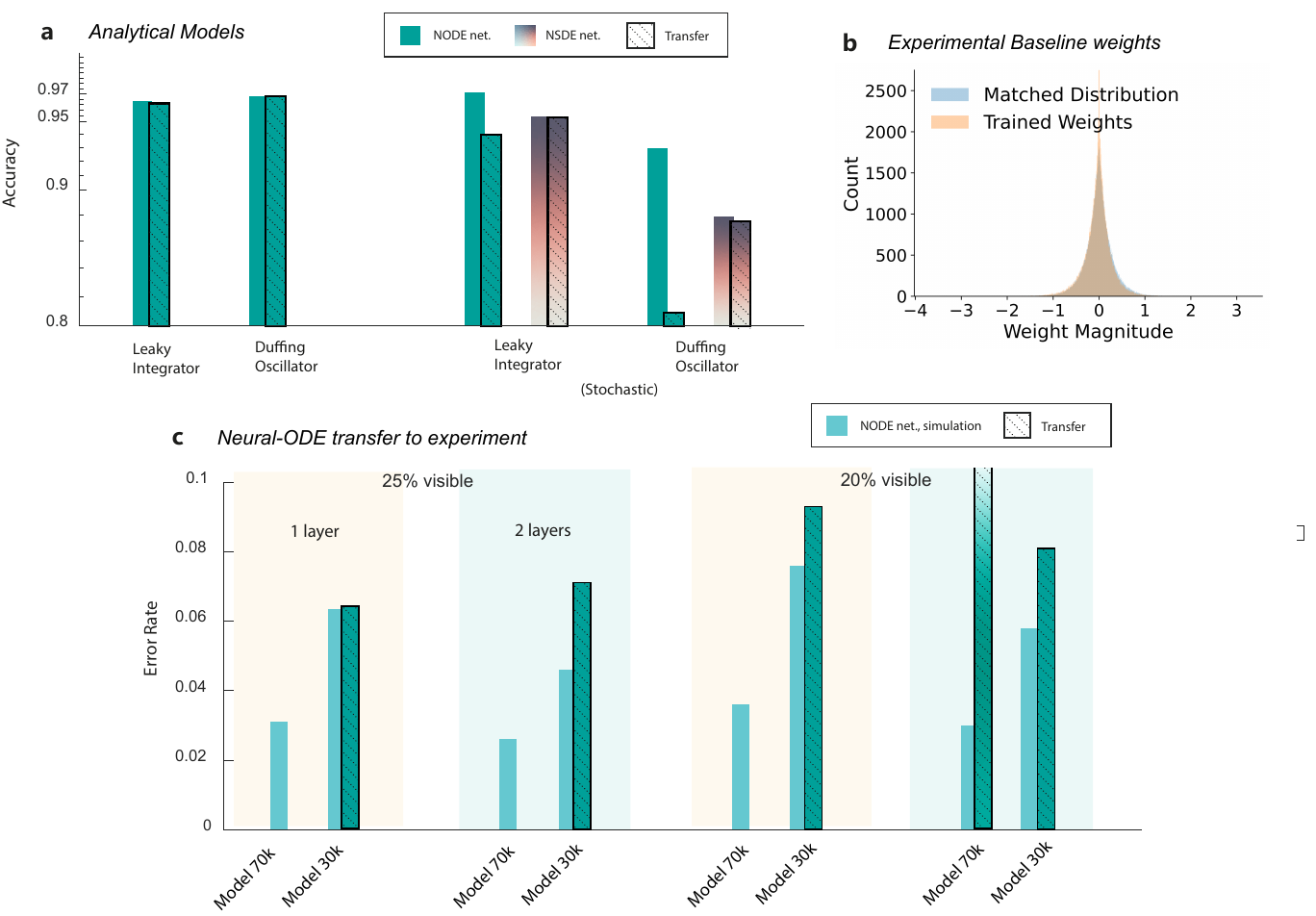}
\caption{\textbf{a} Reference performance on the MNIST variation for two-layer networks of the analytical models. While the Neural-SDEs are accurate predictor of the systems behaviours and lead to optimised parameters that are robust with respect to stochasticity, the neural-ODEs fail short for the noisy variants of the systems. \textbf{b} Example distributions of input weights taken from (orange) a multi-layer perceptron trained via the digital twin approach, and (blue) a parameter-matched Laplace distribution used for the extreme learning machine frameworks. \textbf{c} Performance of a network of digital twins composed by neural-ODEs in simulation and after transferring to the experiment. The models 70k and 30k refer to networks trained for $3 \times 10^{5}$ and $7 10^{5}$ learning updates with a batch size of $50$. While further optimisation lead to improved performance in simulation, parameters transferred from the Model 70k were unable to obtain a meaningful classification. This is showcased for the two-layer network and $25\%$ visibility. The result can be interpreted in the following way: as optimisation progresses, the parameters become more dependent on the deterministic responses provided by the neural-ODE, tending to amplify the simulation-reality gap. For this reason, we had to stop the optimisation process before convergence (at $3 \times 10^{5}$) for acceptable performance with transferred parameters. As shown in the main text, the noise-aware optimisation of the neural-SDE removes this difficulty.  }
\label{fig:Supp_perfBaseline}
\end{figure*}

Figure \ref{fig:Supp_perfBaseline}(b) shows the performance of neural-ODE and neural-SDE for the two analytical systems under consideration, including their deterministic and stochastic variants, and two-layer network structures. As observed in Figure \ref{fig:Supp_perfGrids}, the digital twins of Neural-ODEs are in excellent agreement with the deterministic variants of the systems. For these results on the Duffing Oscillator, we allowed inputs to be in range $[-0.1 \ 0.1]$ to avoid bifurcations. 

Networks trained via Neural-ODEs exhibit a notable loss in performance when transferred to stochastic systems (right side of Figure \ref{fig:Supp_perfBaseline} a). In contrast, the digital twins of Neural-SDEs more accurately predicted performance of the true networks, maintaining accuracy when transferring connectivity due to the noise-aware training conducted by the stochastic models. For this stochastic variant of the Duffing Oscillator, we included noise as specified in the above section and allowed inputs over the mentioned $[-0.2 \ 0.2]$ range to enable noise-induced bifurcations in its dynamics. 
\subsubsection*{Extreme Learning Machine of Ring Array Networks for the MNIST benchmark}
To establish a performance baseline for random connectivity structures for the partially observable MNIST task, we exploited the framework of extreme learning machines (ELM)\cite{ding2014extreme}. Here, the dynamics of the nanoring array are used as a nonlinear temporal kernel, but with randomly sampled weights. The distribution of the random input weights was determined by matching the mean and variance of a Laplace distribution such that the sampled weights resemble those of a trained network, shown in Figure \ref{fig:Supp_perfBaseline} b. In this way, the relative magnitudes of weighted connections are comparable, but the specific weights between input channels are not maintained. This process is repeated for connectivities between additional hidden layers. The final output weights are then trained via backpropagation using a binary cross entropy loss function, identically to the digital twin framework. Since the input/hidden weights are fixed after random sampling, the dynamic properties are not considered, hence backpropagation through time is not required, similar to the reservoir computing framework\cite{jaeger_echo_2001}. The reduction in performance observed in the ELM networks highlights the importance of optimisation and the consequent dependencies among specific weight values.
\subsubsection*{Extreme Learning Machines for Neuroprosthetics Task.}
\begin{figure*}[h!]
\centering
\includegraphics[width=0.9\columnwidth]{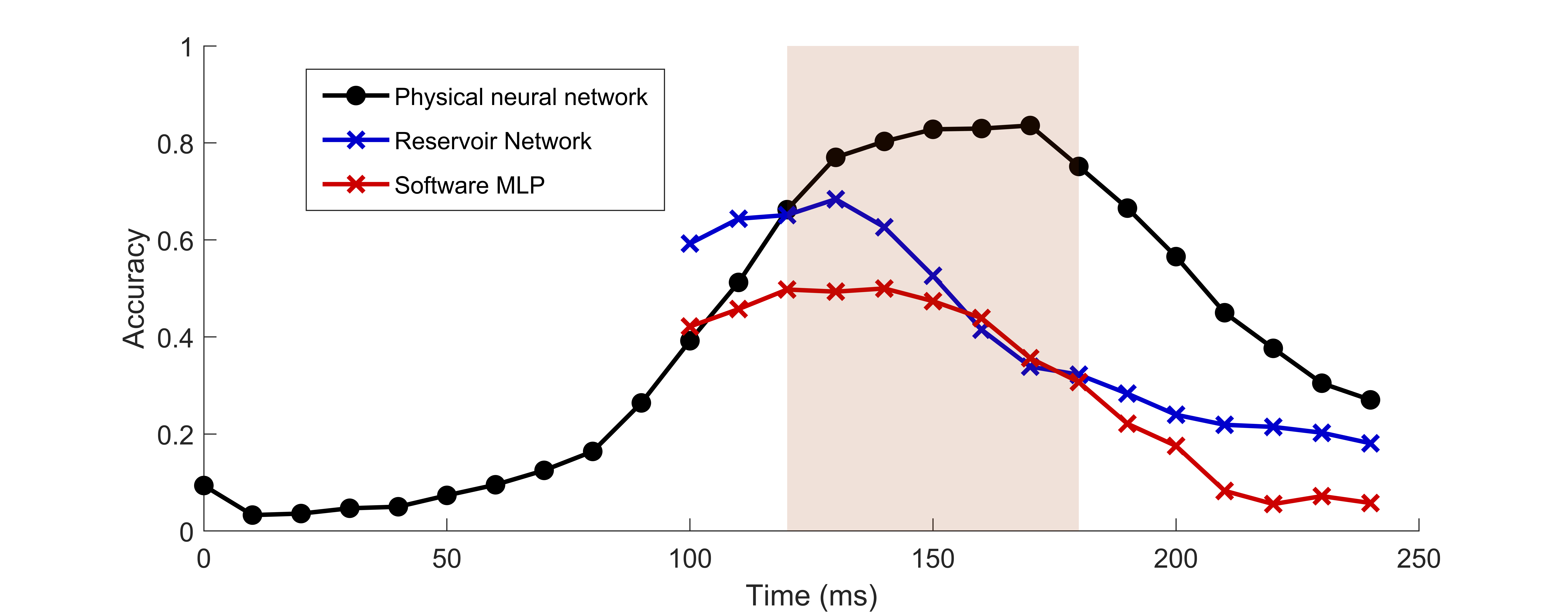}
\caption{Experimental and reference performance on the neuroprosthetic task as the classification time varies, for optimised networks of nanoring arrays (black), random connectivities of nanoring arrays (red), and software neural networks with small memory buffers (blue). Training is performed over a window containing the most meaningful .}
\label{fig:Supp_pNeuro}
\end{figure*}

Due to the inability to provide meaningful connections between hidden layers highlighted by the sharp decrease in performance with two-layer ELMs in the partially observable MNIST task, the baseline performance for the Neuroprosthetics task adopted a single-layer ELM. The width of the layer of 400 nodes has been chosen to match the two hidden layers of the optimised dynamic networks of devices. The mean and distribution of random input weights were optimised via a grid search according to the highest performance on a validation data set. The training was performed over the same window resembling meaningful information as was used in Figure \ref{fig:Fig3}. Similarly, we show the performance of a software MLP of identical shape to the dynamic networks and sigmoidal activation functions. To introduce a finite memory source similar to that provided by the nanoring devices, the inputs instead consisted of the previous three entries for each of the input dimensions, creating a buffer of previous inputs. Since the network itself has no dynamic properties, standard backpropagation was performed using the same binary cross-entropy loss function as in the dynamic networks. Figure \ref{fig:Supp_pNeuro} shows performances obtained when classifying at each timestep of the input signals for a single trained model optimised over the highlighted window, showing the optimised dynamic networks achieve greater classification accuracy maintained over a longer duration.

\begin{figure*}[h!]
\centering
\includegraphics[width=0.85\columnwidth]{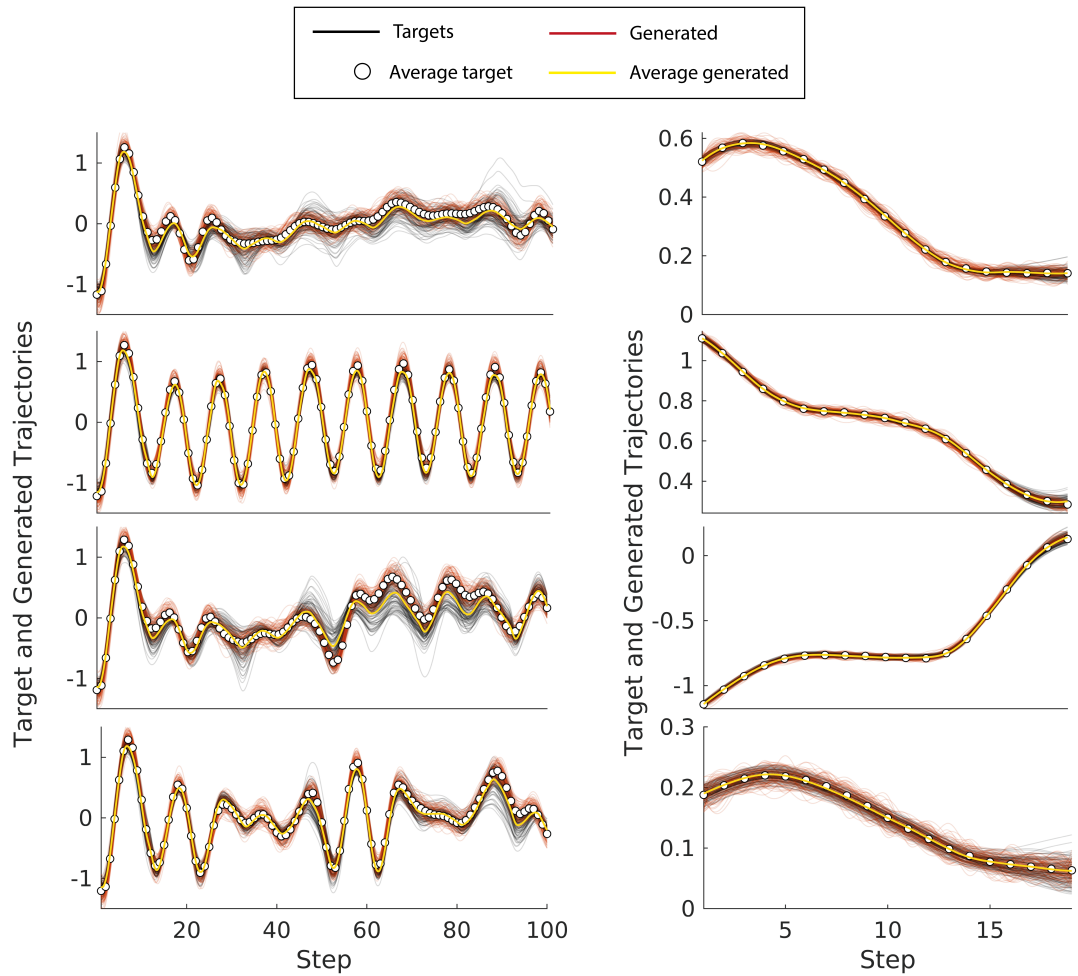}
\caption{Comparison between generated and target dynamics for the physical systems considered, NRA (left) and ASVI (right). In each panel, the systems have been driven by an external signal $\mathbf{s}(t)$, while different panels correspond to different external signals. Similarly to Figure \ref{fig:Fig_Sup_analytical}, the trajectories generated by the neural-SDE are in black, while the reference trajectories are depicted in red. The right panels corresponding to the ASVI illustrate the dynamics of different frequencies.  }
\label{fig:Fig_Exp_Traj}
\end{figure*}
\end{document}